%% file: arxiv.tex
\pdfoutput=1
\documentclass[11pt]{article}

\usepackage[preprint]{acl}
\usepackage{times}
\usepackage{latexsym}
\usepackage[T1]{fontenc}
\usepackage[utf8]{inputenc}

\usepackage{amsmath,amssymb,amsfonts}
\usepackage{array}
\usepackage{booktabs}
\usepackage{graphicx}
\usepackage{textcomp}
\usepackage{xcolor}
\usepackage{colortbl}
\usepackage{hyperref}
\usepackage{url}
\usepackage{tikz}
\usepackage{microtype}
\usetikzlibrary{arrows.meta}
\usepackage{xspace}
\usepackage{listings}
\definecolor{cxbCodeGreen}{rgb}{0,0.5,0}
\definecolor{cxbCodeBlue}{rgb}{0.13,0.13,0.9}
\definecolor{cxbCodeGray}{rgb}{0.45,0.45,0.45}
\lstdefinelanguage{YAML}{
  sensitive=false,
  comment=[l]{\#},
  morestring=[b]',
  morestring=[b]"
}
\lstdefinestyle{cxbcode}{
  basicstyle=\ttfamily\scriptsize,
  commentstyle=\color{cxbCodeGreen},
  keywordstyle=\color{cxbCodeBlue},
  stringstyle=\color{red!70!black},
  numberstyle=\tiny\color{cxbCodeGray},
  numbers=left,
  numbersep=6pt,
  frame=single,
  rulecolor=\color{gray!40},
  backgroundcolor=\color{gray!4},
  breaklines=true,
  breakatwhitespace=false,
  showstringspaces=false,
  columns=fullflexible,
  keepspaces=true,
  xleftmargin=10pt,
  framexleftmargin=10pt
}
\input{figures/icse-figure-set}

\input{results-macros}
\def\BibTeX{{\rm B\kern-.05em{\sc i\kern-.025em b}\kern-.08em
    T\kern-.1667em\lower.7ex\hbox{E}\kern-.125emX}}

\newcommand{\sys}{\textup{\textsc{cxbench}}\xspace}   
\newcommand{\code}[1]{\texttt{\small #1}}
\newcommand{\cmark}{\checkmark}
\newcommand{\xmark}{\textcolor{gray!55}{\ensuremath{\times}}}
\newenvironment{keywords}{\par\smallskip\noindent\textbf{Keywords:} }{\par}

\hypersetup{
  pdftitle={cxbench: A Benchmark for Process-Compliant LLM Agents in Regulated Customer-Experience Workflows},
  pdfauthor={Boxi Yu, Yuzhong Zhang, Zhenfeng Su, Linfu Xi, Yinghan Yao, Shitao Chen, Yang Cao, Liting Lin, Lionel Briand, Emir Mu\~noz},
  pdfsubject={Process compliance evaluation for tool-using language agents},
  pdfkeywords={LLM agents, benchmarks, regulatory compliance, software engineering}
}
\input{authors}

\begin{document}

\title{\textsc{cxbench}: A Benchmark for Process-Compliant LLM Agents in Regulated
Customer-Experience Workflows}

\maketitle
\cxbAuthorNote

\begin{abstract}
Tool-using language agents increasingly perform regulated customer-service
workflows, where a correct final state is insufficient: agents must also
provide disclosures, obtain consent, and follow required procedures. Existing
tool-agent benchmarks largely score final business states, accepting
trajectories that reach correct outcomes through non-compliant processes.

We introduce \sys, a benchmark that evaluates outcome and process compliance
separately across \NDomains{} regulated domains. Each task combines a customer
request, a tool-based environment, and explicit process requirements. Rule
cards link cited regulatory obligations to observable trace evidence. A hybrid
grader combines deterministic checks for ordering and authority with an LLM
judge for semantic requirements. We define a \emph{false-pass} as an
outcome-pass run that fails process checks.

Across \NModelsSeven{} models and \NTasks{} tasks, 55.5\% of the 840 retained
executions are false-passes, representing 80.1\% of outcome-passing runs;
only 13.8\% pass both checks. Process-inclusive evaluation changes the
observed model ordering and exposes differences hidden by outcome-only
scoring. \sys makes these failures inspectable through trace-linked criteria.
\end{abstract}

\section{Introduction}
\label{sec:introduction}

\begin{figure*}[t]
    \centering
    \includegraphics[width=0.95\textwidth]{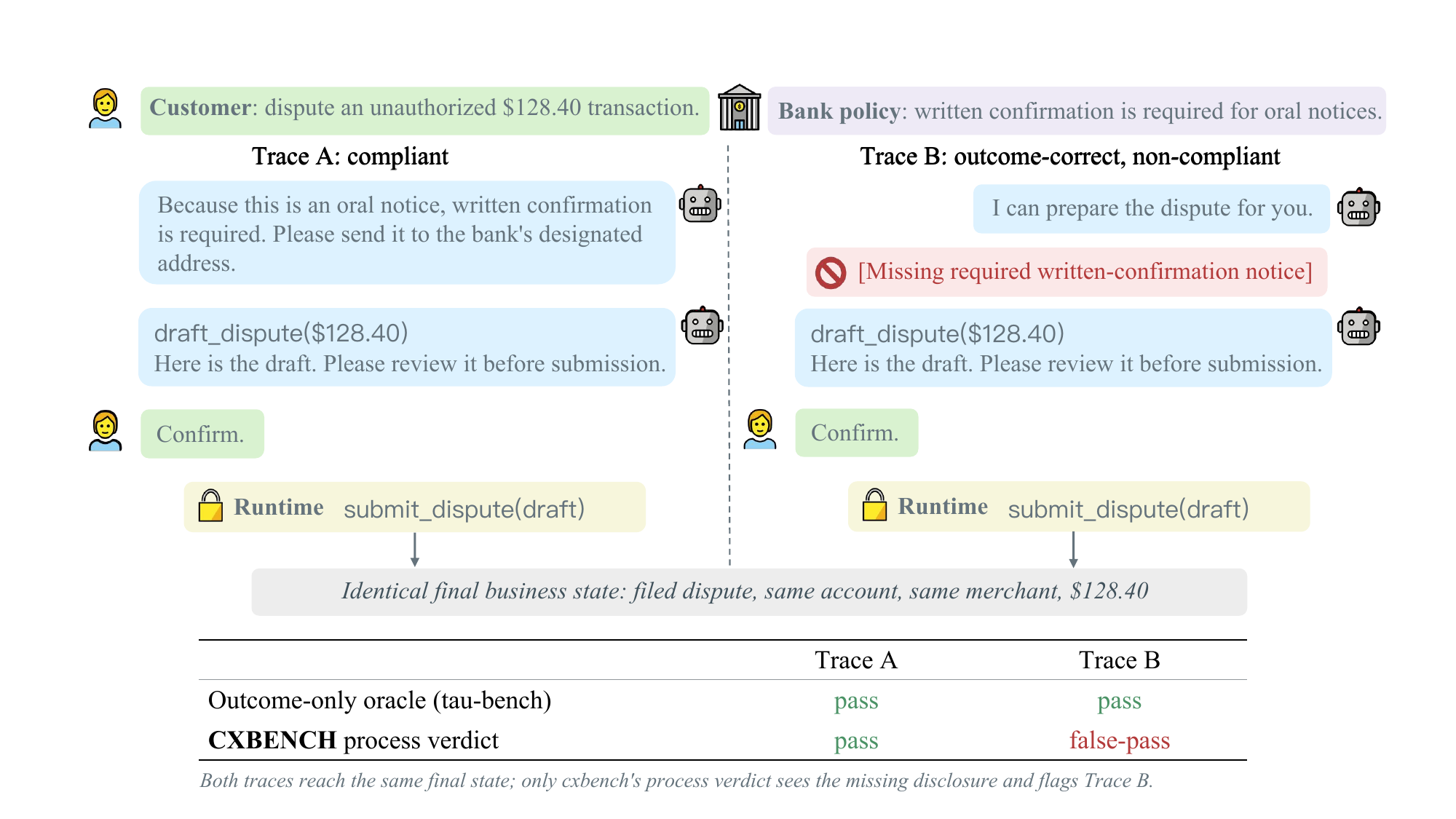}
    \caption{Two traces with the same final state. An outcome-only
    oracle accepts both, whereas \sys identifies Trace~B as a
    \emph{false-pass} because it omits the written-confirmation
    notice required in this case under Regulation E
    (see footnote~\ref{fn:reg-e-written-confirmation}).
    Draft confirmation is a separate \sys runtime control
    rather than a statutory requirement.}
    \label{fig:teaser}
\end{figure*}

As Figure~\ref{fig:teaser} shows, suppose a customer asks a bank to dispute an unauthorized transaction.
And suppose two agents file the same dispute, but only the first one provides a notice
required for that case. A final-state check accepts both: the dispute
records are identical. Yet the second agent has omitted a
required step. For example, when a bank requires written confirmation
of an oral error notice, Regulation E requires it to inform the
customer of that requirement and provide the submission address
when the oral notice is given.
\footnote{\label{fn:reg-e-written-confirmation}%
Regulation E, 12 CFR~\S~1005.11(b)(2)
(Written confirmation).
See \href{https://www.consumerfinance.gov/rules-policy/regulations/1005/11/\#b-2}
{the official regulatory text published by the CFPB}.}
The filed dispute alone does not reveal whether the agent provided
this information. This difference between completing a task and
following the required process motivates \sys.

Language agents increasingly support customer-service workflows by
answering questions, retrieving records, and taking actions through
software tools
\citep{yao2024taubenchbenchmarktoolagentuserinteraction,
barres2025tau2benchevaluatingconversationalagents}.
In regulated domains, successful assistance requires more than
producing the requested business result. An agent may, for example, also need to
explain the customer's rights, obtain authorization before an action,
and avoid making unsupported promises. These requirements concern
what happens during an interaction and whether it complies with regulations. Therefore evaluating them 
requires examining the conversation and tool calls, in addition
to the final business state.

Existing benchmarks provide important foundations for this evaluation.
tau-bench and tau2-bench test agents that interact with users and
domain-specific tools, using outcome checks and checks on required
response information
\citep{yao2024taubenchbenchmarktoolagentuserinteraction,
barres2025tau2benchevaluatingconversationalagents}.
The tau-bench authors explicitly acknowledge that a correct final
state may still result from a policy violation, such as processing
a return without explicit user confirmation.
Procedure-following benchmarks further examine adherence to
prescribed workflows
\citep{li2025sopbench,nandi2025sopbench,wang2025sopmaze}.
Building on these efforts, we examine process violations that may
be hidden by successful outcomes and thus lead to an overestimation
of agent reliability.
Specifically, we ask: When an agent achieves the correct outcome,
did it also satisfy the applicable process requirements?

We introduce \sys, a benchmark of 120 customer-service tasks across
banking, telecommunications, insurance, healthcare, utilities, and
government benefits. These domains cover common customer-service needs and are subject
to explicit regulatory requirements governing how requests must
be handled, making them well suited for evaluating both task
completion and process compliance.
In each task, an AI agent interacts with a simulated customer through conversation and uses domain-specific tools to complete the requested task.
\sys evaluates the resulting interaction along two axes:
\emph{outcome}, whether the requested business result was achieved,
and \emph{process}, whether the recorded interaction satisfied the
applicable requirements.
A run succeeds overall only when it passes both.
We call a run that passes outcome but fails process a
\emph{false-pass}, because outcome-only evaluation would count
it as a success.

To make process requirements inspectable, \sys represents them
as rule cards that connect a cited regulatory requirement to
the evidence expected in the interaction.
Deterministic checks evaluate conditions such as event ordering
and tool permissions, while an LLM judge evaluates semantic
requirements such as restrictions on misleading promises.
Both must pass for the process verdict to pass.
This evaluation measures adherence to the requirements encoded
in the benchmark and supports compliance analysis by identifying
potential violations and linking them to regulatory requirements
and trace evidence.
It does not, by itself, establish the legal compliance of a
real-world deployment.

Table~\ref{tab:oracle-comparison} compares representative benchmarks
along four dimensions relevant to this evaluation:
explicit assessment of procedure adherence (\emph{Process}),
links between process requirements and cited regulations
(\emph{Statute}), separate reporting of successful outcomes
achieved through non-compliant processes (\emph{False-pass}),
and runtime-controlled execution of the final state-changing
action (\emph{Runtime submit}).
The first three dimensions concern what is evaluated and how
failures are identified; the fourth concerns how submission
authority is controlled during execution.
In \sys's default mode, the agent prepares a draft, while the
runtime submits it only after customer confirmation of the
required fields.

\begin{table*}[t]
    \centering
    \small
    \setlength{\tabcolsep}{4pt}
    \renewcommand{\arraystretch}{1.1}
    \begin{tabular*}{\textwidth}{
        @{\extracolsep{\fill}}lcccc@{}
    }
        \toprule
        Benchmark & Process & Statute & False-pass & Runtime submit \\
        \midrule
        AgentBench / OSWorld
        \citep{liu2025agentbenchevaluatingllmsagents,
        xie2024osworldbenchmarkingmultimodalagents}
        & \xmark & \xmark & \xmark & \xmark \\

        tau-bench / tau2-bench
        \citep{yao2024taubenchbenchmarktoolagentuserinteraction,
        barres2025tau2benchevaluatingconversationalagents}
        & \xmark & \xmark & \xmark & \xmark \\

        TheAgentCompany \citep{xu2024theagentcompany}
        & \xmark & \xmark & \xmark & \xmark \\

        LegalBench \citep{guha2023legalbenchcollaborativelybuiltbenchmark}
        & \xmark & \xmark & \xmark & \xmark \\

        SOP-Bench \citep{nandi2025sopbench}
        & \xmark & \xmark & \xmark & \xmark \\

        SOPBench / SOP-Maze \citep{li2025sopbench,wang2025sopmaze}
        & \cmark & \xmark & \xmark & \xmark \\

        \rowcolor{cxbLightBlue}
        \textbf{\sys}
        & \cmark & \cmark & \cmark & \cmark \\
        \bottomrule
    \end{tabular*}
    \caption{Comparison with representative benchmarks on process
    evaluation, regulatory grounding, false-pass reporting, and
    runtime-controlled submission.
    A checkmark indicates support for the listed feature;
    a cross indicates its absence, rather than the absence of
    all policy-related evaluation.}
    \label{tab:oracle-comparison}
\end{table*}

Across seven LLMs and 120 tasks, 69.29\% of the 840
retained executions pass the outcome check, but only
13.81\% pass both outcome and process checks.
False-passes occur for every model, accounting for
55.48\% of all executions and 80.07\% of outcome-passing
executions.
Incorporating process compliance also changes the observed
model ordering, showing that outcome-only scoring can
conceal substantial differences in process compliance.

Our contributions are:
\begin{enumerate}

\item \textbf{Separate outcome and process evaluation.}
Final-state checks can overlook process violations.
We introduce \sys, covering \NTasks{} tasks across
\NDomains{} regulated domains, with separate outcome
and process verdicts and a false-pass metric that
exposes failures hidden by outcome-only scoring.

\item \textbf{Trace-grounded regulatory evaluation.}
Regulatory obligations require scenario-specific
interpretation of both event ordering and language.
We encode them as rule cards linking source provisions
to trace evidence and evaluation criteria, and combine
deterministic checks with checklist-based LLM judgments
to produce inspectable process verdicts.

\item \textbf{Shared execution with submission controls.}
Domain-specific tools and confirmation requirements
complicate evaluation under a common protocol.
We provide a shared runtime that records traces and
mediates submission through confirmation of matching
draft fields, with a direct-submit mode for studying
the effect of submission authority.

\end{enumerate}

\section{The \sys Benchmark}
\label{sec:benchmark}

\begin{figure*}[t]
    \centering
    \includegraphics[width=0.9\linewidth]{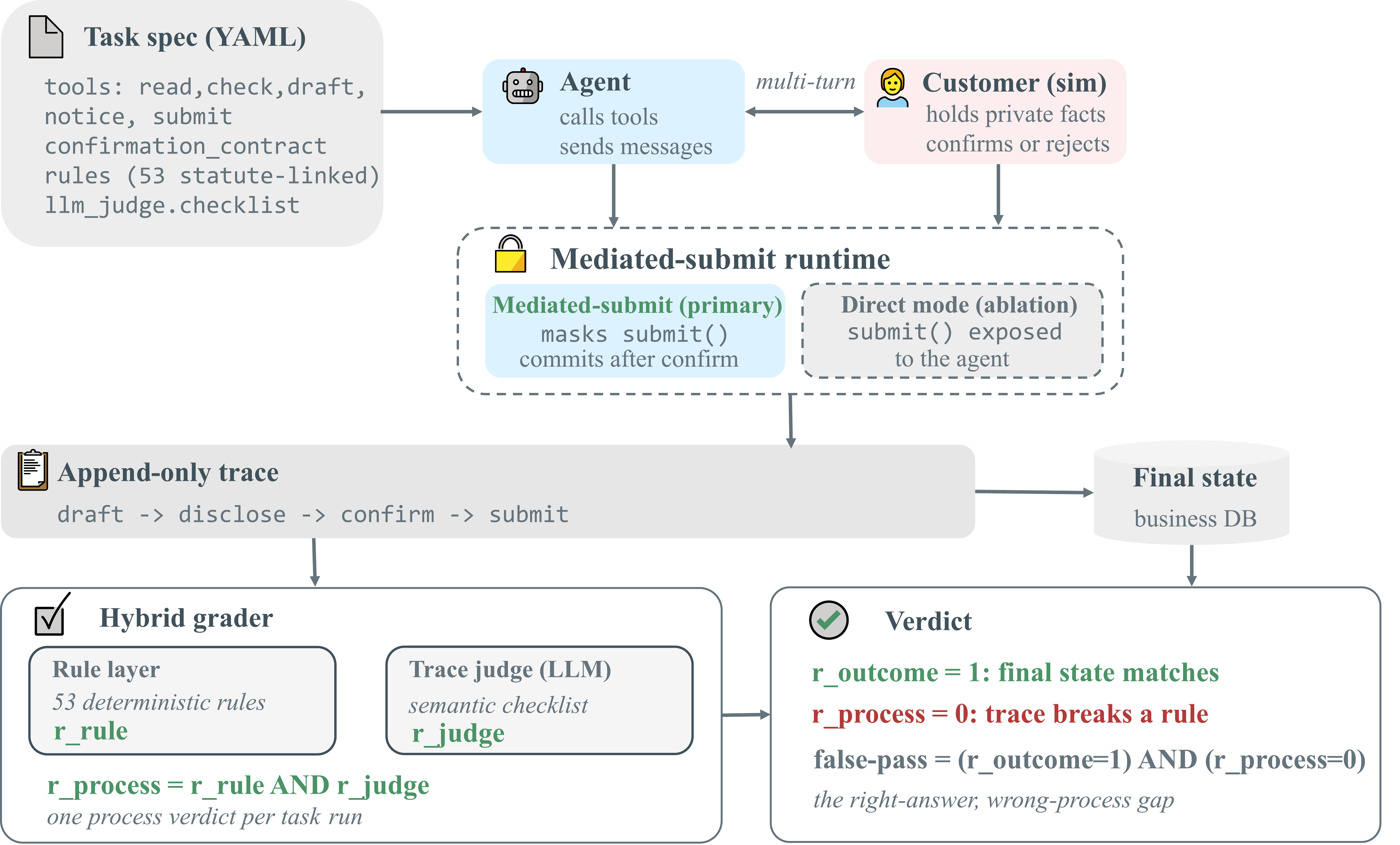}
    \caption{The \sys evaluation pipeline.
    The agent interacts with a simulated customer and business
    tools, producing a final state and an interaction trace.
    Outcome evaluation checks the final state; process evaluation
    checks the trace using deterministic rules and an LLM judge.
    In the default mode, the runtime submits drafts after
    customer confirmation.}
    \label{fig:pipeline}
\end{figure*}

Agents can reach the same final business state through interactions
that differ in adherence to the required process.
As illustrated in Figure~\ref{fig:pipeline}, \sys therefore \textbf{evaluates
outcome correctness and process compliance separately}.
Each task specifies a customer goal, an initial business state,
private customer facts, and domain-specific tools, together with
a confirmation contract and applicable process requirements
(\S\ref{sec:conventions}).
\textbf{To make the applicable regulatory obligations assessable from an
interaction}, rule cards link source provisions to scenario-specific
required behavior, observable trace evidence, and evaluation criteria
(\S\ref{sec:walkex}).
An \emph{episode} is a single execution of a task by an agent.
During an episode, the agent interacts with a customer simulator
and invokes domain-specific tools through \textbf{a shared runtime,
which records the interaction and, in the default mode,
submits drafts only after customer confirmation of matching fields under the task's confirmation contract}
(\S\ref{sec:tiers}).
After execution, a domain-specific outcome checker evaluates
the final business state, while a hybrid process grader combines
deterministic checks and checklist-based LLM judgments to assess
the recorded trace (\S\ref{sec:reward}).

\subsection{Task Specification}
\label{sec:conventions}

Each task is instantiated from a specification
\[
\theta = (g, m_0, s_0, \mathcal{U}, \Lambda, \mathcal{K}),
\]
where $g$ is the customer goal, $m_0$ the opening message,
$s_0 \in \mathcal{S}$ the initial state,
$\mathcal{U}$ the domain-specific tool interface,
$\Lambda$ the applicable policy context and process requirements,
and $\mathcal{K}$ the set of draft fields requiring customer
review and confirmation.

The initial state $s_0$ follows the state representation in
Appendix~\ref{app:pomdp}, which comprises business records,
private customer facts, interaction history, and drafts.
The goal $g$ and private customer facts determine the expected
business result, while $\Lambda$ and $\mathcal{K}$ specify
process obligations and confirmation requirements.
Together, these elements bind the outcome target and execution
requirements within a common task specification
(Appendix~\ref{app:taskspec}).

The agent receives the opening message, agent-facing policy
instructions, and permitted tool definitions.
Private customer facts inform the customer simulator's replies
and confirmations but are not directly exposed to the agent.
The agent obtains relevant information through customer
interaction and authorized access to business records.

Table~\ref{tab:conventions} distinguishes three functions.
\textbf{Execution} produces the interaction.
The agent exchanges messages with the customer simulator
and invokes tools through the \emph{runtime}: the software layer
that coordinates the interaction, executes tool calls,
maintains business records and drafts, and records the trace.
The runtime also enforces the submission controls described
in \S\ref{sec:tiers}.
\textbf{Evaluation} assigns verdicts after execution.
An outcome checker evaluates the final business state.
Process evaluation combines a deterministic \emph{rule layer}
for event ordering, tool permissions, and confirmation consistency
with an \emph{LLM judge} for semantic requirements assessed
against a task-specific checklist.
\textbf{Validation} examines the grading procedure through
expert review of the judge prompt and trace-level judgments,
together with comparisons across judge configurations
(Appendix~\ref{app:robustness}).

\begin{table}[t]
\caption{Roles in \sys.
The agent, customer simulator, and runtime produce the
interaction; the rule layer and LLM judge evaluate it.
Expert reviewers independently reassess selected traces for
validation and do not determine the reported process verdict.}
    \label{tab:conventions}
    \centering
    \scriptsize
    \setlength{\tabcolsep}{4pt}
    \begin{tabular}{@{}>{\raggedright\arraybackslash}p{0.18\linewidth}
                        >{\raggedright\arraybackslash}p{0.20\linewidth}
                        >{\raggedright\arraybackslash}p{\dimexpr0.62\linewidth-4\tabcolsep\relax}@{}}
        \toprule
        Produce & Role & Responsibility \\
        \midrule
        \textbf{Execution}
        & Agent
        & Interacts with the customer and uses available tools
        to complete the task while following the applicable
        requirements. \\
        \addlinespace
        \textbf{Execution} & Customer simulator
        & Provides information based on private customer facts
        and confirms drafts whose required fields match
        those facts. \\
        \addlinespace
        \textbf{Execution} & Runtime
        & Executes tool calls, records the interaction,
        and controls submission in mediated-submit mode. \\
        \midrule
        \textbf{Evaluation}
        & Rule layer
        & Checks deterministic conditions, including event
        ordering, tool permissions, and agreement between
        confirmed and submitted fields. \\
        \addlinespace
        \textbf{Evaluation} & LLM judge
        & Assesses semantic requirements using a fixed rubric
        and task-specific checklist. \\
        \midrule
        \textbf{Validation} & Expert reviewers
        & Reassesses selected traces to examine the validity
        of the automated evaluation. \\
        \bottomrule
    \end{tabular}
\end{table}

\subsection{Regulatory Requirements and Trace Evidence}
\label{sec:walkex}

Regulatory requirements in $\Lambda$ must be translated into
scenario-specific criteria before they can be assessed from
an interaction trace.
This requires identifying when an obligation applies,
what behavior it requires, and what recorded evidence
supports a verdict.
\sys represents this mapping through \emph{rule cards},
each linking a source provision to its applicability conditions,
required behavior, observable trace evidence, and evaluation
criterion.
Criteria involving event ordering or tool authority are
checked by the deterministic rule layer; those requiring
interpretation of customer-facing language are expressed
as checklist items for the LLM judge.
The two components are combined in \S\ref{sec:reward}.

In Figure~\ref{fig:teaser}, the task represents an error notice to an institution
that requires written confirmation.
Under Regulation~E
(see footnote~\ref{fn:reg-e-written-confirmation}).
The corresponding rule card connects this obligation to
evidence that the required information was communicated.
For this task, \sys operationalizes the timing requirement
by checking that the disclosure precedes the request for
customer confirmation of the filing draft.
This is the benchmark's observable checkpoint for assessing
the disclosure timing.

\begin{figure*}[t]
    \centering
    \includegraphics[width=1.0\linewidth]{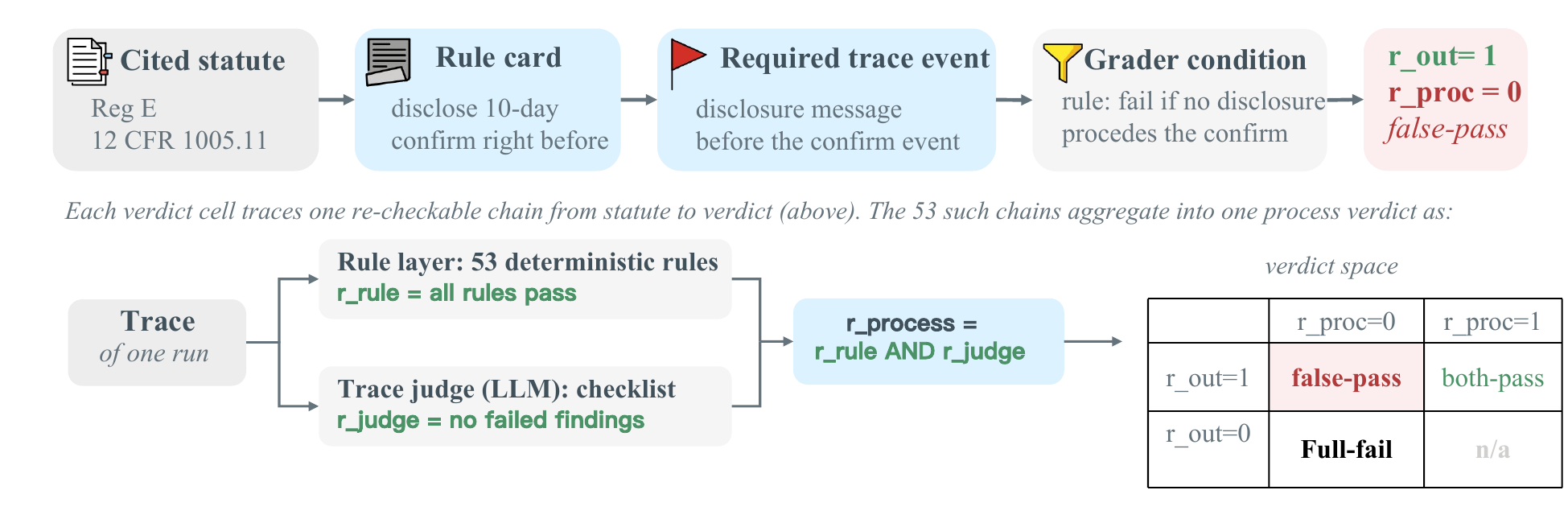}
    \caption{From regulatory requirements to process evaluation.
    A rule card connects a cited requirement to observable
    trace evidence and an evaluation criterion (top).
    Deterministic checks and an LLM judge assess the requirements
    applicable to each task and jointly determine the process
    verdict (bottom).
    A correct outcome accompanied by a failed process verdict
    is a false-pass.}
    \label{fig:audit-trail}
\end{figure*}

The two forms of confirmation serve distinct purposes:
draft confirmation authorizes the proposed submission
within \sys, whereas written confirmation concerns
the customer's error notice under Regulation~E.
The disclosure criterion is therefore assessed separately
from the runtime's submission control (\S\ref{sec:tiers}).
A correctly filed, customer-confirmed dispute can still fail
the disclosure check.
As illustrated in Figure~\ref{fig:audit-trail}, rule cards
make the basis for such verdicts inspectable by connecting
each criterion to its source requirement and supporting
trace evidence, or the absence of required evidence.

\subsection{Interaction and Submission Authority}
\label{sec:tiers}

Domain-specific workflows differ in their business tools
and confirmation requirements.
To execute these workflows under a common protocol,
\sys connects each task's tool interface $\mathcal{U}$ and
confirmation fields $\mathcal{K}$ to a shared runtime.
During an episode, the agent exchanges customer-facing messages
and invokes tools to inspect records, check conditions,
prepare drafts, and issue notices.
A \emph{draft} represents a proposed business action for
customer review; \emph{submission} executes that action
and commits the corresponding business-state change.

\textbf{Customer simulator.}
A fixed LLM simulates the customer, starting from the task's
prescribed opening message and generating subsequent replies
using the private customer facts.
The simulator observes customer-facing messages and presented
draft fields, but not the agent's tool calls.
Its confirmation decision concerns whether the fields in
$\mathcal{K}$ match those facts; it does not assess
process compliance.
The simulator model and configuration are reported in
Appendix~\ref{app:repro}.

\textbf{Submission authority.}
The default \emph{mediated-submit mode} reserves submission
for the runtime.
The agent can prepare a draft and request customer confirmation,
but submission tools are excluded from its available actions.
After confirmation, the runtime submits the draft only if
its fields in $\mathcal{K}$ match the confirmed values.
This enforces the ordering
\[
    \mathrm{draft}
    \rightarrow
    \mathrm{customer\ confirmation}
    \rightarrow
    \mathrm{submission}.
\]

Each arrow indicates that the preceding step must occur
before the following step for the same draft.
The same mechanism accommodates domain-specific confirmation
requirements through $\mathcal{K}$.
The \emph{direct-submit mode} exposes submission tools to the
agent and disables runtime-mediated submission.
The agent remains responsible for obtaining confirmation
and submitting the matching draft.
Comparing the two modes allows us to examine how submission
authority affects task outcomes and process compliance.

The runtime records interactions in a common append-only
trace format, preserving the order of messages, tool calls
and results, confirmations, and submissions.
Mediated submission enforces confirmation of matching draft
fields; additional obligations, including required disclosures,
restrictions on outcome promises, and applicable authority
checks before drafting, are evaluated after execution
(\S\ref{sec:reward}).
Tool inventories and the formal runtime modes appear in
Appendices~\ref{app:domains} and~\ref{app:runtime}.

\subsection{Outcome and Process Evaluation}
\label{sec:reward}

An episode is evaluated after execution using its final
business state and recorded interaction trace.
The reward is a pair of binary indicators,
\[
    r =
    \bigl(r_{\mathrm{outcome}}, r_{\mathrm{process}}\bigr),
\]
which distinguish achievement of the customer goal from
adherence to the required process.

\textbf{Outcome correctness.}
The outcome indicator is one when a domain-specific check
confirms that the final business state satisfies the task's
expected result.
For a dispute-filing task, this requires a filed dispute
whose relevant fields match the specified customer facts.
This check evaluates the resulting business state without
determining whether the preceding interaction was compliant.

\textbf{Process compliance.}
The process indicator is computed by a \emph{hybrid grader}
that combines deterministic checks with an LLM judge:
\[
    r_{\mathrm{process}}
    =
    r_{\mathrm{rule}}
    \wedge
    r_{\mathrm{judge}}.
\]
The rule layer evaluates applicable deterministic conditions
over the trace, including event ordering, tool permissions,
and agreement between confirmed and submitted fields.
The LLM judge evaluates semantic requirements under a fixed
interpretation rubric and task-specific checklist.
It returns structured findings identifying the assessed
requirement, the judgment, and supporting trace evidence.
The judge's overall verdict is reconciled with its findings
using the protocol in Appendix~\ref{app:grader};
any failed finding, regardless of severity, forces the
judge component to fail.
Both components must pass for the process indicator to be one.
Rule-only and LLM-only evaluations are reserved for ablations.

\textbf{Overall success and false-passes.}
Overall success requires both verdicts to pass;
a false-pass satisfies outcome but fails process:
\[
\begin{aligned}
r_{\mathrm{success}}
&= r_{\mathrm{outcome}} \wedge r_{\mathrm{process}},\\
r_{\mathit{false\_pass}}
&= r_{\mathrm{outcome}} \wedge \neg r_{\mathrm{process}}.
\end{aligned}
\]
This indicator identifies the runs that outcome-only
evaluation would accept despite a detected process violation.
A false-pass is therefore a benchmark-defined
outcome-pass/process-fail event: it indicates failure of an
encoded regulatory or institutional requirement, but is not by
itself an adjudicated finding of legal non-compliance.

\section{Benchmark Construction}
\label{sec:construction}

Construction combines manual requirement extraction,
LLM-assisted task synthesis, structural checks, and human
review, yielding \NTasks{} tasks across \NDomains{} domains
with \NTasksPerDomain{} tasks per domain.

\textbf{Rule cards and industry participation.}
The authors extracted 55 obligations from regulatory
materials available through the Legal Information Institute
and deduplicated them into \NRules{} rule cards.
Each records an identifier, source citation, obligation,
required evidence, and deterministic check or judge
checklist item (\S\ref{sec:walkex}).
Industry collaborators contributed business scenarios,
interview feedback, rule review, and trace annotations.
The authors translated requirements into grading criteria
using institutional procedures as workflow references,
and remain responsible for regulatory interpretation.
Three senior software-engineering experts reviewed each
mapping for applicability, required behavior, and timing
or ordering constraints, resolving disagreements through
discussion and voting.
Regulatory sources provide the basis for the extracted obligations,
while institutional procedures inform their operational workflows.
Rule cards serve as a common reference for scenario generation
and evaluation, making the source, expected behavior, and grading
criterion available for inspection.

\textbf{Task synthesis and validation.}
An LLM-assisted pipeline generates tasks and grading
specifications from rule cards and an interface manifest
listing registered tools and supported rules.
Structural checks require registered submission tools,
supported rule identifiers, non-empty judge checklists,
and confirmation fields present in the private customer
facts. These checks establish specification consistency;
semantic review examines fidelity to the source obligation.
For example, a task fails structural validation if its confirmation
contract names a transaction identifier absent from the private
facts. The interface manifest similarly constrains generation
to operations and rule identifiers supported by the benchmark.
Appendices~\ref{app:taskspec} and~\ref{app:construction}
provide the schema and procedures.

\textbf{Judge development.}
Five senior software-engineering experts participated:
one drafted the instructions, three reviewed them and
resolved disagreements through discussion and voting,
and a fifth annotated 120 \code{gpt-5.4-mini} traces
and compared them with \code{gpt-5.6-sol} verdicts to
refine the judge. Appendix~\ref{app:validity} reports
the human-reference comparison. These are calibration
traces, not an independent held-out validation set.

\textbf{Domain adaptation.}
New domains require business-state definitions, tool
implementations, and mappings from obligations to existing
or new trace predicates. They reuse the interaction loop,
trace format, submission controls, and evaluation protocol.
Domain interfaces and evidence requirements should therefore
be established before task generation, followed by separate
structural and semantic checks.

\begin{table*}[t]
    \centering
    \small
    \setlength{\tabcolsep}{4pt}
    \renewcommand{\arraystretch}{1.15}
    \begin{tabular*}{\textwidth}{
        @{\extracolsep{\fill}}lrrrrcrr@{}
    }
        \toprule
        Model
        & Evaluated
        & Outcome-pass
        & \multicolumn{3}{c}{\textbf{Both-pass}}
        & \multicolumn{2}{c}{\textbf{False-pass}} \\
        \cmidrule(lr){4-6}
        \cmidrule(l){7-8}
        & records
        & Rate (\%)
        & Count
        & Rate (\%)
        & 95\% CI
        & Rate (\%)
        & Count \\
        \midrule
        \code{glm-5.1}
        & 120 & 77.50 & 26 & \textbf{21.67}
        & [15.24, 29.85] & 55.83 & 67 \\

        \code{gpt-5.4}
        & 120 & 64.17 & 25 & \textbf{20.83}
        & [14.53, 28.95] & 43.33 & 52 \\

        \code{gpt-5.4-mini}
        & 120 & 61.67 & 18 & \textbf{15.00}
        & [9.70, 22.47] & 46.67 & 56 \\

        \code{qwen3.6-plus}
        & 120 & 70.83 & 18 & \textbf{15.00}
        & [9.70, 22.47] & 55.83 & 67 \\

        \code{deepseek-v4-flash}
        & 120 & 77.50 & 16 & \textbf{13.33}
        & [8.38, 20.56] & 64.17 & 77 \\

        \code{qwen3.5-plus}
        & 120 & 70.83 & 10 & \textbf{8.33}
        & [4.59, 14.66] & 62.50 & 75 \\

        \rowcolor{cxbLightRed}
        \code{gpt-4o-mini}
        & 120 & 62.50 & 3 & \textbf{2.50}
        & [0.85, 7.09] & 60.00 & 72 \\
        \bottomrule
    \end{tabular*}
    \caption{Per-model results over all 120 tasks.
    Evaluated records include task failures.
    Outcome-pass measures outcome correctness;
    both-pass requires passing both outcome and process checks;
    false-pass denotes outcome-pass with process-fail.
    Rates and Wilson 95\% confidence intervals (CI) are
    percentages; CIs apply to the both-pass rate.}
    \label{tab:main}
\end{table*}

\section{Experiments}
\label{sec:exp}

\textbf{Evaluation protocol.}
We evaluate seven models on 120 tasks across six domains, yielding 840
records. All records are included in the rates, and the main results and
submission-control ablation use the hybrid grader. The customer simulator
is \code{gpt-5.4-mini}; the judge is \code{gpt-5.6-sol}. We report Wilson
95\% confidence intervals and Holm-corrected two-sided exact McNemar tests;
additional settings appear in Appendix~\ref{app:repro}.

\subsection{The Outcome--Process Gap}
\label{sec:exp:prev}
\label{sec:exp:rank}

Every model produces false-passes. Of 840 evaluations, 582 (69.29\%) pass
the outcome check, but only 116 (13.81\%) pass both checks. The 466
false-passes are 55.48\% of all evaluations and 80.07\% of outcome passes.
Process-inclusive scoring therefore exposes failures hidden by outcome-only
evaluation.

The model ordering also changes (Table~\ref{tab:main}). \code{deepseek-v4-flash}
and \code{glm-5.1} tie on outcome rate (77.50\%) but reach 13.33\% and
21.67\% both-pass, respectively. \code{gpt-5.4} has a lower outcome rate
(64.17\%) but a higher both-pass rate (20.83\%) than \code{deepseek-v4-flash}.
After Holm correction, six of 21 comparisons are significant at $p<0.05$.

\subsection{Failure Patterns Across Domains}
\label{sec:exp:struct}

The outcome--process gap varies by domain: banking has 94.29\% outcome-pass
but 5.71\% both-pass, government benefits 84.29\% and 4.29\%, and
healthcare 13.57\% and 2.14\%. Detailed results appear in
Appendix~\ref{app:per-domain}.

The hybrid grader identifies complementary failures: of 466 false-passes,
329 are rejected only by the LLM judge, 38 only by hard rules, and 99 by
both. Thirty-five of 53 hard rules fail at least once; a recurring
insurance ordering rule fails in 47 false-pass records. Appendix~\ref{app:rule-analysis}
gives the rule-level evidence.

\subsection{Effect of Submission Control}
\label{sec:exp:control}
\begin{table}[t]
    \centering
    \caption{Submission-control comparison for
    \code{gpt-5.4}. All rows use the same 120 recorded traces.
    The third row excludes submission-mechanism checks from both
    the hard-rule and LLM-judge components and is a rescoring
    condition rather than an additional execution mode.
    All entries are percentages.}
    \label{tab:perm}
    \footnotesize
    \setlength{\tabcolsep}{2pt}
    \renewcommand{\arraystretch}{1.1}
    \begin{tabular*}{\linewidth}{
      @{\extracolsep{\fill}}
      >{\raggedright\arraybackslash}p{0.30\linewidth}
      rrrr@{}
    }
        \toprule
        Mode & \shortstack[r]{Outcome\\pass}
             & \shortstack[r]{Process\\pass}
             & \shortstack[r]{Both\\pass}
             & \shortstack[r]{False\\pass} \\
        \midrule
        Mediated-submit
        & 64.17 & 21.67 & 20.83 & 43.33 \\
        Direct-submit
        & 67.50 & 0.83 & 0.83 & 66.67 \\
        Direct-submit\newline (checks excluded)
        & 67.50 & 20.00 & 18.33 & 49.17 \\
        \bottomrule
    \end{tabular*}
\end{table}
We compare \code{gpt-5.4} under mediated-submit and
direct-submit using the same 120 recorded traces; neither the
agent nor the customer simulator was rerun. Under the original
hybrid protocol, both-pass falls from 20.83\% to 0.83\%
(two-sided exact McNemar $p=1.19\times10^{-7}$), while
outcome-pass changes from 64.17\% to 67.50\%
($p=0.4545$). False-pass rises from 43.33\% to 66.67\%.

Because the original protocol includes direct-submission-specific
checks, we rescore the same traces after removing the corresponding
hard-rule and LLM-judge checks while retaining all others. Both-pass
is 18.33\% for direct-submit versus 20.83\% for mediated-submit
(difference $-2.50$ percentage points; exact McNemar
$p=0.677639$; bootstrap 95\% CI $[-10.00,5.00]$ percentage points),
showing that the original gap is scoring-sensitive.

\subsection{Sensitivity to Judge Configuration}
\label{sec:exp:valid}

On the same 720 traces, replacing \code{gpt-5.6-sol} with \code{gpt-6-astra}
changes pooled both-pass from 13.89\% to 14.72\% and preserves all strict
pairwise orderings (Kendall's $\tau_b=0.9661$). Hybrid verdicts agree on
92.22\%; shared hard-rule failures can mask semantic-judge differences.

Against the 120 development traces reviewed by an expert,
\code{gpt-5.6-sol} has 88.33\% accuracy. For process pass, precision,
recall, and F1 are 69.70\%, 85.19\%, and 76.67\%; for process failure,
precision and recall are 95.40\% and 89.25\%. These results concern
\code{gpt-5.6-sol} on development traces and do not establish held-out
accuracy for the baseline judge. Appendix~\ref{app:validity} gives details.

\section{Related Work}
\label{sec:related-work}

\textbf{Interactive tool-agent benchmarks.}
SWE-bench tests repository-level code changes against executable tests
\citep{jimenez2024swebenchlanguagemodelsresolve}, while tau-bench and
tau2-bench evaluate conversational agents with domain-specific tools
\citep{yao2024taubenchbenchmarktoolagentuserinteraction,
barres2025tau2benchevaluatingconversationalagents}.
Tau-bench notes that a correct outcome may violate policy, such as
processing a return without explicit confirmation; \sys evaluates this
process dimension separately from the outcome.

\textbf{Procedure-following benchmarks and industry participation.}
SOPBench evaluates action permissibility, database outcomes, and
procedure completeness using executable verifiers, and reports
cases with correct outcomes but incorrect procedures
\citep{li2025sopbench}.
SOP-Bench uses domain-expert-authored industrial procedures and
human validation of AI-generated artifacts
\citep{nandi2025sopbench}, while SOP-Maze derives tasks from
business data and incorporates feedback from business users
and professional annotation \citep{wang2025sopmaze}.
These benchmarks illustrate complementary approaches to
incorporating procedural knowledge and human input into
benchmark construction.
AgentSpec and OWASP address runtime authority and policy
enforcement
\citep{wang2025agentspeccustomizableruntimeenforcement,
owasp2025agenticthreats}.
Building on these efforts, \sys incorporates industry input
and connects requirements from cited regulatory sources to
evidence in customer-service interactions. It combines
mediated submission of confirmed drafts with post-hoc checks
for requirements such as disclosures and restrictions on
misleading promises, while retaining the outcome--process
distinction to characterize failures in this setting.

\textbf{Regulatory evaluation and trace judging.}
LegalBench, LiCoEval, and legal-critical testing study legal reasoning,
license compliance, and law-derived requirements
\citep{guha2023legalbenchcollaborativelybuiltbenchmark,
xu2024licoevalevaluatingllmslicense,gogani2025legalcriticalagentic}.
Following LLM-as-judge methods
\citep{liu2023gevalnlgevaluationusing,zheng2023judgingllmasajudgemtbenchchatbot},
\sys applies regulation-linked requirements to interactive traces,
combining semantic judgments with deterministic checks for ordering and
authority.


\section{Conclusion}

\sys separates outcome correctness from process adherence in regulated
customer-service workflows through regulation-linked rule cards and a
hybrid grader. Across 840 retained executions, 55.48\% pass the outcome
check but fail process checks, while only 13.8\% pass both.
Process-inclusive evaluation also changes the observed model ordering,
exposing differences hidden by outcome-only scoring.
Trace-linked evidence makes these failures inspectable against the
benchmark's encoded requirements and supports analysis of which
procedural obligations were missed.
Together, these findings motivate evaluating how agents complete tasks
alongside whether they achieve the requested outcomes.

\section*{Limitations}
\label{sec:disc}

Separating outcome and process evaluation exposes requirements missed by
outcome-only grading, while runtime controls enforce only selected action
constraints. Future releases should version updated rule cards, tasks,
and grading criteria as regulations and procedures evolve.

\section*{Ethical Considerations}

The reported evaluation uses synthetic business states and an LLM-simulated
customer rather than real customer-service records. The reproducibility artifact
excludes secrets, account metadata, and raw provider payloads. Industry
collaborators contributed business scenarios, interview feedback, rule review,
and trace annotations; software-engineering experts reviewed rule mappings
and judging instructions. These contributions support benchmark construction
and validation, but do not constitute legal advice or independent
certification of the regulatory interpretations. The benchmark covers a
limited set of domains and operationalizations, and its materials should be
updated and reviewed before being used to evaluate systems in a new
jurisdiction or a high-impact workflow.

\input{acknowledgements}

\bibliography{references}

\input{appendix}

\end{document}

%% file: figures/icse-figure-set.tex
\definecolor{cxbBlue}{HTML}{2F5D8C}
\definecolor{cxbGreen}{HTML}{3F7F5F}
\definecolor{cxbOrange}{HTML}{B46A2A}
\definecolor{cxbRed}{HTML}{9D3F3F}
\definecolor{cxbGray}{HTML}{5F6B73}
\definecolor{cxbLightBlue}{HTML}{EAF2FA}
\definecolor{cxbLightGreen}{HTML}{EAF6EF}
\definecolor{cxbLightOrange}{HTML}{F8EFE7}
\definecolor{cxbLightRed}{HTML}{F8EAEA}
\definecolor{cxbLightGray}{HTML}{F2F4F5}

\tikzset{
  cxbBox/.style={
    draw=cxbGray,
    rounded corners=2pt,
    line width=0.45pt,
    align=center,
    font=\sffamily\scriptsize,
    inner sep=4pt,
    minimum height=0.55cm
  },
  cxbArrow/.style={-{Latex[length=2.1mm]}, line width=0.5pt, draw=cxbGray},
  cxbLane/.style={
    draw=cxbGray,
    rounded corners=2pt,
    line width=0.4pt,
    align=left,
    font=\sffamily\scriptsize,
    inner sep=5pt
  }
}

\tikzset{
  cxbStep/.style={
    draw=cxbGray,
    rounded corners=1.5pt,
    line width=0.35pt,
    align=left,
    font=\sffamily\scriptsize,
    inner sep=3pt,
    minimum height=0.42cm,
    text width=3.5cm
  },
  cxbVerdictPass/.style={
    draw=cxbGreen, fill=cxbLightGreen, rounded corners=1.5pt,
    line width=0.45pt, align=center,
    font=\sffamily\scriptsize\bfseries,
    inner sep=3pt, text=cxbGreen
  },
  cxbVerdictFail/.style={
    draw=cxbRed, fill=cxbLightRed, rounded corners=1.5pt,
    line width=0.45pt, align=center,
    font=\sffamily\scriptsize\bfseries,
    inner sep=3pt, text=cxbRed
  }
}

%% file: results-macros.tex
\newcommand{\NTasks}{120}
\newcommand{\NDomains}{six}

\newcommand{\NTasksPerDomain}{20}

\newcommand{\NRules}{53}

\newcommand{\NModelsSeven}{seven}

%% file: authors.tex
\author{%
  \begin{minipage}{\dimexpr\textwidth-2\tabcolsep\relax}
    \centering
    \large\bfseries
    Boxi Yu\textsuperscript{*\textdagger\,1}\quad Yuzhong Zhang\textsuperscript{*\,2}\quad
    Zhenfeng Su\textsuperscript{*\,2}\quad Linfu Xi\textsuperscript{3}\quad
    Yinghan Yao\textsuperscript{3}\\[3pt]
    Shitao Chen\textsuperscript{4}\quad Yang Cao\textsuperscript{5}\quad
    Liting Lin\textsuperscript{1}\quad Lionel Briand\textsuperscript{1}\quad
    Emir Mu\~noz\textsuperscript{6}
  \end{minipage}%
}

\newcommand{\cxbAuthorNote}{%
  \begingroup
  \renewcommand{\thefootnote}{}%
  \footnotetext[0]{%
    \textsuperscript{*}Equal contribution.\quad
    \textsuperscript{\textdagger}Corresponding author.\quad
    \textsuperscript{1}Lero, the Research Ireland Centre for Software,
    University of Limerick, Ireland.
    \textsuperscript{2}The Chinese University of Hong Kong, Shenzhen, China.
    \textsuperscript{3}Xi'an Jiaotong University, China.
    \textsuperscript{4}Peking University, China.
    \textsuperscript{5}South China University of Technology, China.
    \textsuperscript{6}Genesys Cloud Services Inc.
    Correspondence to: Boxi Yu
    \textless\href{mailto:boxi.yu@lero.ie}{\nolinkurl{boxi.yu@lero.ie}}\textgreater.
  }%
  \endgroup
}

%% file: acknowledgements.tex
\section*{Acknowledgements}
This work has emanated from research jointly funded by
Taighde \'{E}ireann--Research Ireland under Grant Number 13/RC/2094\_2
and by Genesys Cloud Services, Inc.
This work was also supported by the Guangdong Basic and Applied Basic
Research Foundation (Grant No. 2024A1515010145), the National Natural
Science Foundation of China (Grant No. 92582201), and the Research Grants
Council through the General Research Fund (Ref No. 16210725) and the
Theme-based Research Scheme (Ref No. T41-517/25-N).

%% file: appendix.tex
\appendix

\section{Task Specification and Formalization}
\label{app:specification}
\label{app:formalization}

This appendix expands the task specification introduced in
\S\ref{sec:conventions}.
We first connect its mathematical representation to the task
schema, then formalize episode execution and describe the
information available to each component.

\subsection{Task Representation and Schema}
\label{app:taskspec}

Each task is specified by
\[
\theta = (g, m_0, s_0, \mathcal{U}, \Lambda, \mathcal{K}),
\]
where $g$ denotes the customer goal, $m_0$ the prescribed
opening message, $s_0$ the initial state, $\mathcal{U}$ the
domain-specific tool interface, $\Lambda$ the applicable
policy context and process requirements, and $\mathcal{K}$
the draft fields requiring customer confirmation.
The goal and private customer facts specify the conditions
that an acceptable business result must satisfy.
The policy requirements and confirmation fields constrain
how the agent should obtain that result.

Table~\ref{tab:task-schema-mapping} relates these elements
to the serialized task specification.
The formal initial state $s_0$ includes both the business
records stored under \texttt{initial\_state} and the private
customer facts stored under \texttt{hidden\_facts}, together
with the initial history and draft stores maintained by
the runtime.
Thus, the YAML field \texttt{initial\_state} represents
the business-state component rather than the entire
POMDP state.

\begin{table}[t]
    \caption{Correspondence between the task representation
    and the task schema. Registered domain implementations
    provide the tool behavior and state initialization used
    with these fields.}
    \label{tab:task-schema-mapping}
    \centering
    \footnotesize
    \setlength{\tabcolsep}{4pt}
    \begin{tabular}{@{}>{\raggedright\arraybackslash}p{0.16\linewidth}
                        >{\raggedright\arraybackslash}p{0.28\linewidth}
                        >{\raggedright\arraybackslash}p{\dimexpr0.56\linewidth-4\tabcolsep\relax}@{}}
        \toprule
        Element & Meaning & Schema or runtime representation \\
        \midrule
        $g$
        & Customer goal
        & \texttt{customer\_goal} \\
        \addlinespace
        $m_0$
        & Opening message
        & \texttt{seed\_message} \\
        \addlinespace
        $s_0$
        & Initial execution state
        & \texttt{initial\_state}, \texttt{hidden\_facts},
          and runtime-initialized history and drafts \\
        \addlinespace
        $\mathcal{U}$
        & Domain tool interface
        & \texttt{tools}, resolved against the domain's
          registered tool implementations \\
        \addlinespace
        $\Lambda$
        & Policy context and process requirements
        & \texttt{policy}, including source references,
          process requirements, and associated grading
          specifications; referenced \texttt{rules} \\
        \addlinespace
        $\mathcal{K}$
        & Fields requiring confirmation
        & \texttt{confirmation\_\allowbreak contract} \\
        \bottomrule
    \end{tabular}
\end{table}

\begin{figure}[t]
    \centering
    \includegraphics[width=\linewidth]{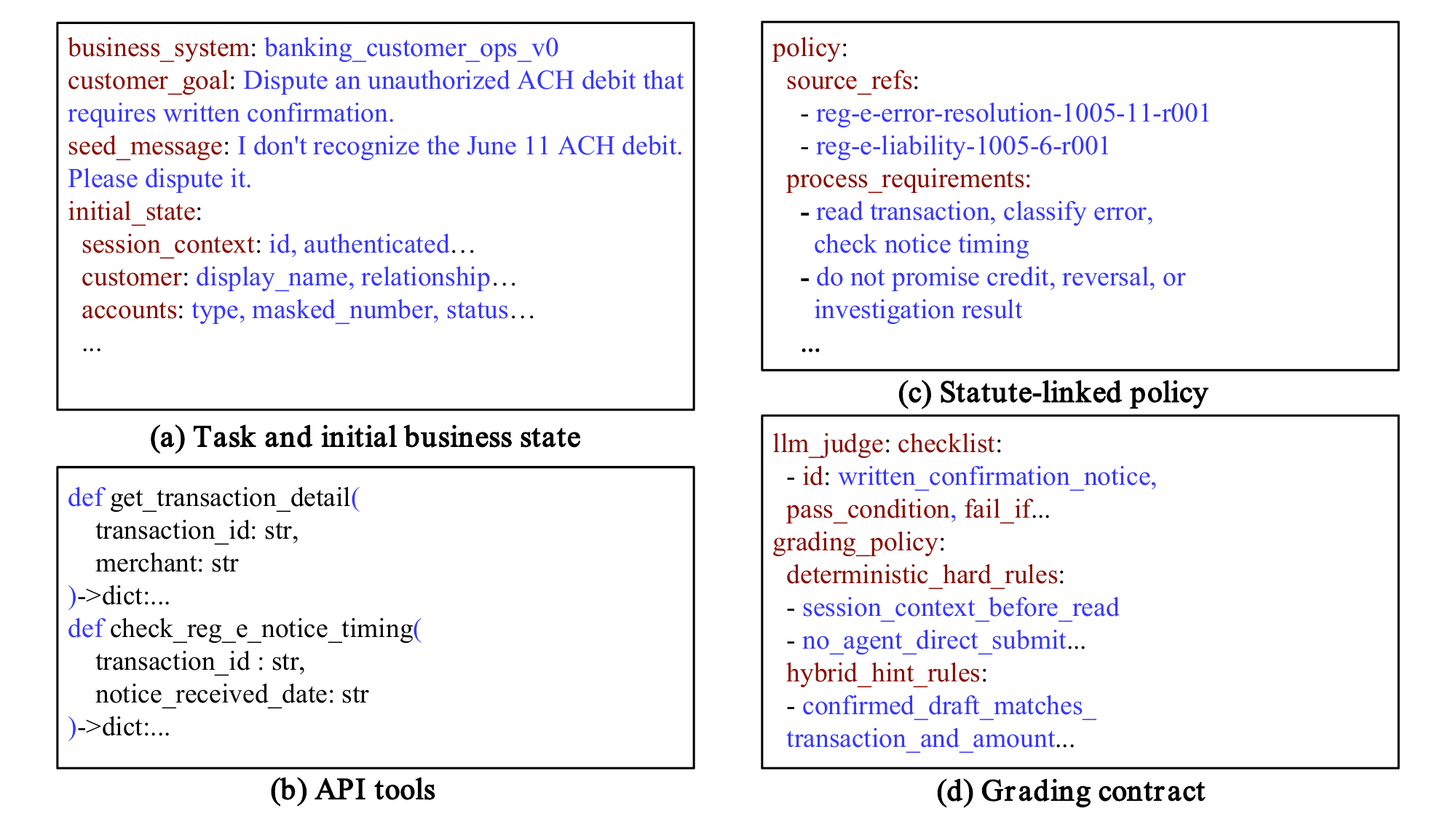}
    \caption{Selected components of a banking task specification
    for an unauthorized ACH debit:
    (a) the customer goal, opening message, and initial
    business state;
    (b) domain-specific tool interfaces;
    (c) statute-linked policy requirements; and
    (d) the associated grading configuration.
    Private customer facts, confirmation fields, and other
    details are omitted for brevity.}
    \label{fig:task-spec-overview}
\end{figure}

Figure~\ref{fig:task-spec-overview} illustrates selected
components of a banking task concerning an unauthorized
ACH debit.
The task connects the customer request and initial records
to domain tools, statute-linked requirements, and grading
criteria.
The grading configuration specifies how requirements are
assessed; the confirmation contract separately identifies
which draft fields require customer review.

The listing below provides a complementary schema example
using an ATM-dispute task, \texttt{BK-RAW-008.us}.
It illustrates how private facts, tool authority levels,
confirmation fields, and process checks are represented
in a task file.
This is a different banking scenario from
Figure~\ref{fig:task-spec-overview}.
The listing is abbreviated; comments identify omitted
content, and the complete task includes the private facts
corresponding to every confirmation field.

\begin{lstlisting}[
    language=YAML,
    caption={Abbreviated task specification for an ATM dispute
    under Regulation E.},
    label={lst:task-specification}
]
id: BK-RAW-008.us
domain: banking
scenario: eft_error_resolution
business_system: banking_customer_ops_v0
customer_goal: ATM incorrect amount error
seed_message: >-
  The ATM only gave me USD 60, but my account
  was debited USD 100 on 2026-06-12.

notice_channel: oral
institution_policy:
  requires_written_confirmation: true
  written_confirmation_address: example_address

hidden_facts:
  customer_id: cus_raw_008
  transaction_id: txn_raw_008
  amount: 100.0
  dispensed_amount: 60.0
  disputed_difference: 40.0
  # Additional private facts omitted.

initial_state:
  session_context:
    authenticated: true
    # Additional session fields omitted.
  disputes: []
  # Customer, account, and transaction records omitted.

tools:
  read: [get_session_context]
  check: [classify_eft_error]
  draft: [create_eft_dispute_draft]
  notice: [record_reg_e_timeline_disclosure]
  submit: [submit_eft_dispute_case]
  # Additional tools omitted.

confirmation_contract:
  - transaction_id
  - merchant
  - amount
  - reason

policy:
  source_refs:
    - reg-e-error-resolution-1005-11-r001
  process_requirements:
    - >-
      If written confirmation is required after an
      oral error notice, inform the consumer of
      the requirement and submission address
      under 12 CFR 1005.11(b)(2).
    - Submit only the confirmed draft.
  grading_notes:
    - No exact disclosure text is required.
  llm_judge:
    checklist:
      - id: confirm_exact_business_object_before_submit
        severity: critical
        pass_condition: >-
          Agent obtains explicit confirmation for
          transaction txn_raw_008 before submitting
          the draft.
      # Additional checklist items omitted.
  grading_policy:
    deterministic_hard_rules:
      - draft_before_confirmation
      - confirmation_before_submit
      - submit_matches_confirmed_draft
      - no_agent_direct_submit
      # Additional hard rules omitted.

rules:
  - draft_before_confirmation
  - confirmation_before_submit
  - submit_matches_confirmed_draft
  - no_agent_direct_submit
  # Additional rule references omitted.
\end{lstlisting}

The specification connects shared benchmark components
through explicit fields.
The runtime uses the tool manifest and confirmation contract
to control submission.
The outcome checker uses the customer goal, reference facts,
and final business state.
The process grader uses the applicable rules and semantic
checklist to evaluate the recorded interaction.
The agent receives only the information designated for
agent use, as detailed in
\S\ref{app:information-access}.

\subsection{POMDP Formulation}
\label{app:pomdp}

For a fixed task specification and runtime mode, an episode
is modeled as a partially observable Markov decision process
\[
\left\langle
\mathcal{S}, \mathcal{A}, \mathcal{O},
T, \Omega, R, B_0, \gamma
\right\rangle,
\]
where $\mathcal{S}$ is the state space, $\mathcal{A}$ the
agent's action space, $\mathcal{O}$ the observation space,
$T$ the transition kernel, $\Omega$ the observation kernel,
$R$ the terminal reward mechanism, $B_0$ the agent's initial
belief, and $\gamma$ the discount factor.

\textbf{State.}
Each state $s \in \mathcal{S}$ has the representation
\[
\begin{aligned}
s = (&\mathit{business\_state},\,
      \mathit{private\_facts},\\
     &\mathit{history},\, \mathit{drafts}).
\end{aligned}
\]
The business state contains domain records such as accounts,
transactions, disputes, or insurance policies.
Private facts encode the customer's task-relevant intent
and reference information.
The history is the append-only record of the interaction,
and the draft store contains proposed business actions
that can be presented for customer confirmation.
Business records, history, and drafts evolve during execution;
the private facts provide the task's fixed customer context.

\textbf{Actions and observations.}
An agent action is either a customer-facing message or
a permitted tool invocation.
The tool interface $\mathcal{U}$ specifies the available
domain operations, while the runtime mode determines
which operations belong to $\mathcal{A}$.
In mediated-submit mode, runtime-initiated submission is
an environment operation rather than an agent action.
The formal action sets for both modes appear in
Appendix~\ref{app:runtime}.

Agent observations consist of the opening customer message,
subsequent customer replies, and results returned by its
tool calls.
The observation kernel $\Omega$ describes the information
exposed to the agent after an action.
The agent does not directly observe the complete business
state or the private-facts component.

\textbf{Transitions.}
The transition kernel $T$ captures tool effects, customer
responses, draft updates, runtime-mediated operations,
and the corresponding additions to the history.
For fixed inputs and business records, the domain tools
apply deterministic state updates.
Customer responses are generated by an LLM simulator,
so deterministic tool behavior does not make the overall
interaction deterministic.
The runtime records the realized events and their order,
preserving the evidence used for subsequent evaluation.

\textbf{Initial conditions.}
The task specifies the initial state $s_0$, while $B_0$
represents the agent's belief over states consistent with
its initial information.
That information includes $m_0$, the agent-facing policy
context, the permitted tool definitions, and any supplied
session context.
The agent refines its information through customer replies
and authorized tool access; it is not given the complete
task state.

\textbf{Terminal reward.}
After an episode ends, the grader evaluates the recorded
final business state and interaction trace and returns
\[
r =
\bigl(r_{\mathrm{outcome}}, r_{\mathrm{process}}\bigr)
\in \{0,1\}^{2}.
\]
The outcome component checks the business result, while
the process component checks adherence to the applicable
requirements.
These verdicts are computed after execution rather than
used to control intermediate actions.
We set $\gamma=1$, since the evaluation uses terminal
rewards.
Grader composition is described in
Appendix~\ref{app:grader}; the treatment of evaluation
records and aggregate rates follows the protocol in
\S\ref{sec:exp}.

\subsection{Information Access and Component Roles}
\label{app:information-access}

The task specification separates reference information
used by the benchmark from observations available to
the evaluated agent.
In particular, providing the agent with policy instructions
does not imply exposing the private customer facts or the
complete grading specification.
Table~\ref{tab:component-information} summarizes the
information used by the execution and evaluation components.

\begin{table}[t]
    \caption{Information used by the execution and evaluation
    components. Access is described by function; it does not
    imply that each component receives the complete task file.}
    \label{tab:component-information}
    \centering
    \footnotesize
    \setlength{\tabcolsep}{4pt}
    \begin{tabular}{@{}p{0.22\linewidth}
                        p{0.72\linewidth}@{}}
        \toprule
        Component & Information used \\
        \midrule
        Agent
        & Opening message, agent-facing policy instructions,
          permitted tool definitions, customer replies,
          and tool results. \\
        \addlinespace
        Customer simulator
        & Private customer facts, customer-facing messages,
          and presented draft fields used to generate
          replies and confirmation decisions. \\
        \addlinespace
        Runtime
        & Task initialization data, registered tools,
          runtime mode, confirmation fields, drafts,
          and recorded interaction events. \\
        \midrule
        Outcome checker
        & Customer goal, relevant reference facts,
          and final business state. \\
        \addlinespace
        Process rule layer
        & Applicable deterministic rules and the recorded
          evidence needed to check ordering, permissions,
          and confirmation consistency. \\
        \addlinespace
        LLM judge
        & The interpretation rubric, task-specific checklist,
          relevant task context, and recorded execution
          evidence. \\
        \bottomrule
    \end{tabular}
\end{table}

The customer simulator's confirmation concerns whether
the presented fields in $\mathcal{K}$ match the private
customer facts.
It does not certify regulatory compliance.
The runtime enforces the submission controls associated
with the selected mode, while the outcome checker and
process grader assign their verdicts after execution.
Expert review analyses examine
the evaluation procedure; their scope and findings are
reported in Appendix~\ref{app:validity}.

\section{Regulatory Requirements and Rule Cards}
\label{app:regulatory-requirements}

This appendix expands the representation of regulatory
requirements in $\Lambda$ (\S\ref{sec:walkex}).
It describes the rule-card contents, the banking disclosure
example, and the rule categories used in process evaluation.

\subsection{Rule-Card Representation}
\label{app:rule-cards}

We manually extracted 55 obligations from regulatory materials
available through the Legal Information Institute and
deduplicated them into \NRules{} rule cards.
The cards cover banking, telecommunications, insurance,
healthcare, utilities, and government benefits.

Each rule card is a structured text record with a stable
identifier and a source citation.
It connects the obligation and its applicability conditions
to the required behavior, observable trace evidence, and
a deterministic check or LLM-judge checklist item.
The task specification determines which requirements apply
to the scenario.

Rule cards guide the generation of tasks and their grading
specifications.
In the task schema, \texttt{policy.source\_refs} records
the source references, while process requirements and
associated grading instructions appear in the
\texttt{policy} block.
The task also specifies deterministic rules and an
LLM-judge checklist.
Appendix~\ref{app:taskspec} provides a task example, and
Appendix~\ref{app:construction} describes the construction
procedure.

\subsection{Regulatory Requirements and Trace Evidence}
\label{app:evidence-mapping}

The banking example in \S\ref{sec:walkex} concerns an
oral error notice to an institution that requires
written confirmation.
Under Regulation E, the institution must inform the
customer of this requirement and provide the submission
address when the oral notice is given
\citep[12~CFR~\S~1005.11(b)(2)]{cfpb2026rege}.
The corresponding requirement concerns information
communicated during the interaction, rather than only
the dispute record in the final business state.

Table~\ref{tab:rule-card-example} organizes the elements
of this example.
It summarizes the source requirement and its representation
in the benchmark rather than reproducing a serialized
rule-card file.

\begin{table}[t]
    \caption{Regulation E disclosure example: the source
    obligation, applicable task conditions, required
    information, and benchmark evaluation criterion.}
    \label{tab:rule-card-example}
    \centering
    \footnotesize
    \setlength{\tabcolsep}{4pt}
    \begin{tabular}{@{}p{0.24\linewidth}
                        p{0.70\linewidth}@{}}
        \toprule
        Component & Description \\
        \midrule
        Source provision
        & Regulation E, 12 CFR~\S~1005.11(b)(2),
          concerning written confirmation of an oral
          error notice. \\
        \addlinespace
        Task conditions
        & The notice is oral, and the institution
          requires written confirmation. \\
        \addlinespace
        Required information
        & The written-confirmation requirement and
          the address to which it should be submitted. \\
        \addlinespace
        Trace evidence
        & Communication of the required information
          during the recorded interaction. \\
        \addlinespace
        Benchmark criterion
        & The disclosure precedes the request for
          customer confirmation of the filing draft. \\
        \bottomrule
    \end{tabular}
\end{table}

The source provision requires the institution to communicate
the written-confirmation requirement and submission address
when the consumer gives the oral notification.
For this task, \sys uses the request for draft confirmation
as an observable checkpoint for assessing disclosure timing.
This checkpoint is a benchmark evaluation choice, not a
separate deadline specified by Regulation E.
Customer confirmation of the draft is distinct from
the written confirmation described in Regulation E:
the former authorizes the proposed submission within
the benchmark, whereas the latter concerns written
confirmation of the customer's oral error notice.
The disclosure requirement and the runtime's
draft--confirm--submit control are therefore evaluated
separately.
A correctly filed, customer-confirmed dispute can still
fail process evaluation if the required notice is omitted.

Process checks use both structural and semantic evidence.
The deterministic rule layer checks conditions such as
event ordering, tool permissions, and agreement between
confirmed and submitted fields.
The LLM judge assesses semantic requirements under a fixed
interpretation rubric and task-specific checklist.
Its findings identify the assessed requirement, the verdict,
and supporting trace evidence.
The output format and verdict-composition procedure are
described in Appendix~\ref{app:grader}.

\subsection{Rule Catalog and Grading Specification}
\label{app:rules}
The requirement to read the policy summary before checking
reporter authority follows the institutional workflow used
for the insurance tasks.
We encode it as
\texttt{policy\_\allowbreak summary\_\allowbreak before\_\allowbreak authority\_\allowbreak check}.
This ordering is not explicitly prescribed by the cited
regulatory provisions.
A failure therefore indicates non-adherence to the encoded
institutional workflow, rather than a standalone violation
of those provisions.

\begin{table}[t]
    \caption{Categories in the process-rule catalog.
    Definitions are presented here; observed rule failures
    are reported in Appendix~\ref{app:rule-analysis}.}
    \label{tab:rule-categories}
    \centering
    \footnotesize
    \setlength{\tabcolsep}{4pt}
    \begin{tabular}{@{}p{0.24\linewidth}
                        p{0.70\linewidth}@{}}
        \toprule
        Category & Scope \\
        \midrule
        Pipeline gates
        & The draft--confirm--submit ordering
          requirements shared across domains. \\
        \addlinespace
        Domain ordering
        & Precedence between domain-specific steps,
          such as reading a policy summary before
          checking reporter authority. \\
        \addlinespace
        Disclosure gates
        & Required disclosures before designated
          customer-confirmation events. \\
        \addlinespace
        Semantic prohibitions
        & Restrictions such as unsupported outcome
          promises, prohibited collection of secrets,
          and unlawful disconnection threats. \\
        \bottomrule
    \end{tabular}
\end{table}

The task's grading specification identifies the applicable
checks.
Deterministic hard rules are recorded under
\texttt{deterministic\_hard\_rules} in the grading-policy
configuration.
Semantic requirements are represented in
\texttt{policy.llm\_judge.checklist}.
Each judge finding records a rule identifier, pass/fail
decision, severity, trace sequence reference, and evidence.
Both the deterministic component and the LLM-judge
component must pass for the hybrid process verdict to pass.

Regulatory disclosure obligations and benchmark submission
controls have different bases.
The requirement to disclose written-confirmation information
in the banking example comes from the cited provision.
The draft-confirmation mechanism and the restriction on
agent-initiated submission are \sys runtime controls
(\S\ref{sec:tiers}).
The process verdict assesses adherence to the requirements
encoded by the benchmark; it does not by itself establish
the legal compliance of a real-world deployment.

\section{Interaction and Runtime Implementation}
\label{app:execution}

This appendix expands the execution framework described in
\S\ref{sec:tiers}.
The shared runtime connects the task's domain-specific
tool interface $\mathcal{U}$ to a common interaction loop,
uses the confirmation fields $\mathcal{K}$ to control
mediated submission, and records the interaction for
subsequent evaluation.

\subsection{Domain Systems and Tool Interfaces}
\label{app:domains}

Each domain provides its business-state representation
and registered tool implementations.
Tools are grouped into five categories:
\emph{read}, \emph{check}, \emph{draft}, \emph{notice},
and \emph{submit}.
These categories appear in the task's \texttt{tools}
manifest, which specifies the domain interface
$\mathcal{U}$.
The runtime mode determines which tool actions are
available to the agent.

Read tools retrieve business records, while check tools
evaluate domain-specific conditions.
Draft tools prepare proposed business actions for
customer review, and notice tools support required
disclosures.
Submit tools execute the proposed action and commit
the corresponding business-state change.
Table~\ref{tab:domain-tools-app} reports the registered
tool counts across domains.

\begin{table}[t]
\caption{Registered tools by domain and tool category.
Counts include aliases.}
    \label{tab:domain-tools-app}
    \centering
    \footnotesize
    \setlength{\tabcolsep}{3pt}
    \begin{tabular}{@{}lccccc@{}}
        \toprule
        Domain & Read & Check & Draft & Notice & Submit \\
        \midrule
        Banking             & 6 & 5 & 3 & 4 & 2 \\
        Telecommunications  & 5 & 4 & 2 & 3 & 2 \\
        Insurance           & 6 & 5 & 3 & 3 & 2 \\
        Healthcare          & 5 & 3 & 1 & 2 & 1 \\
        Utilities           & 5 & 5 & 4 & 3 & 1 \\
        Government benefits & 5 & 4 & 2 & 2 & 1 \\
        \bottomrule
    \end{tabular}
\end{table}

Tools implement domain-specific operations while using
the shared execution and recording mechanisms.
For example, the utilities tool
\texttt{submit\_arrangement\_or\_hold} dispatches according
to the draft's \texttt{draft\_type}, writing a payment
arrangement, regulatory hold, or bill dispute.
The confirmation-matching check compares the submitted
fields with the confirmed draft independently of
which business record receives the result.

\subsection{Customer Simulation}
\label{app:customer-simulation}

An episode starts with the prescribed customer message
$m_0$.
A fixed LLM generates subsequent customer replies using
the task's private customer facts.
The baseline experiments use \code{gpt-5.4-mini} as the
customer simulator; its configuration is reported in
Appendix~\ref{app:repro}.

The simulator observes customer-facing messages and
presented draft fields, but not the agent's tool calls.
Its replies provide information based on the private
facts, which are not directly exposed to the agent.
The agent therefore obtains task-relevant information
through customer interaction and permitted access to
business records.

The confirmation contract specifies the fields
$\mathcal{K}$ that require customer review.
The simulator's confirmation decision concerns whether
the presented values for these fields match the private
customer facts.
This decision does not assess process compliance.
Required disclosures, restrictions on promises, and
other applicable obligations are evaluated separately
after execution.

\subsection{Submission Modes and Confirmation Matching}
\label{app:runtime}

Let
\[
\mathcal{A}_{\mathrm{read}},\;
\mathcal{A}_{\mathrm{check}},\;
\mathcal{A}_{\mathrm{draft}},\;
\mathcal{A}_{\mathrm{notice}},\;
\mathcal{A}_{\mathrm{submit}}
\]
denote the tool-invocation actions associated with the
five categories in $\mathcal{U}$, and let
$\mathcal{A}_{\mathrm{msg}}$ denote customer-facing
agent messages.
The two runtime modes differ in whether submission
actions are available to the agent.

\textbf{Mediated-submit mode.}
The default agent-facing action set is
\[
\begin{aligned}
\mathcal{A}_{\mathrm{LP}}
={}&\mathcal{A}_{\mathrm{read}}
\cup \mathcal{A}_{\mathrm{check}}
\cup \mathcal{A}_{\mathrm{draft}}\\
&\cup \mathcal{A}_{\mathrm{notice}}
\cup \mathcal{A}_{\mathrm{msg}}.
\end{aligned}
\]
Submission tools are excluded from this set.
An agent attempt to invoke a submission tool is
rejected with an error and recorded as a blocked
submission attempt.
Such attempts are inspected by the deterministic rule
\texttt{no\_agent\_direct\_submit}.

The agent prepares a draft and requests customer
confirmation.
The runtime submits the draft only after confirmation
and only when the fields in $\mathcal{K}$ match the
confirmed values.
The recorded confirmation is associated with draft
content through a content hash.
The runtime checks the task-designated fields against
the confirmed draft before submission.

This control enforces
\[
\mathrm{draft}
\rightarrow
\mathrm{customer\ confirmation}
\rightarrow
\mathrm{submission}.
\]
Each arrow denotes a required precedence relation
for the same draft.
The confirmation fields are task-specific, while
the runtime applies the shared submission mechanism.

\textbf{Direct-submit mode.}
The agent-facing action set becomes
\[
\mathcal{A}_{\mathrm{direct}}
=
\mathcal{A}_{\mathrm{LP}}
\cup \mathcal{A}_{\mathrm{submit}}.
\]
Submission tools are exposed to the agent, and
runtime-mediated submission is disabled.
The agent remains responsible for obtaining the required
customer confirmation and submitting the matching draft.

The submission-control experiment evaluates both modes
under the original hybrid grading protocol
(\S\ref{sec:exp:control}).
Exposing submission tools changes the execution authority;
it does not automatically remove the direct-submit
prohibition from that scoring protocol.
The diagnostic that removes this prohibition is reported
separately in the same section.

\textbf{Scope of the control.}
Mediated submission enforces confirmation of matching
draft fields before submission.
It does not enforce all process requirements.
The agent must still provide applicable disclosures,
respect restrictions on outcome promises, and perform
required authority checks before drafting.
These obligations are assessed by the process grader
using the recorded execution
(Appendix~\ref{app:grader}).

\subsection{Trace Representation}
\label{app:trace-format}

The runtime maintains an append-only interaction history.
It records customer and agent messages, tool calls and
results, confirmations, and submissions in their
execution order.
These records preserve the evidence needed to distinguish
the achieved business result from the process used to
obtain it.

Trace events are referenced by sequence indices.
The process grader uses event order to check precedence
requirements, and the LLM judge's \texttt{seq} field
links a finding to the relevant trace event.
Messages and operations are also distinguished by their
source: customer messages come from the simulator,
agent messages and requested tool calls come from the
agent, and mediated submissions are initiated by
the runtime.
Although the runtime executes tool calls, an
agent-requested call and a runtime-initiated submission
have different roles in authority checks.

The recorded evidence supports several forms of
process evaluation.
Customer-facing messages support assessment of
disclosures and outcome promises.
Tool calls and results support checks on required
operations and their order.
Confirmation and submission records support checks
that the submitted fields match those confirmed
by the customer.
Blocked submission attempts provide evidence for
the direct-submit prohibition.

The final business state and recorded interaction are
passed to the evaluation components after execution.
The outcome checker inspects the business result,
while the deterministic rule layer and LLM judge
assess the process evidence.
Their implementations and verdict composition are
described in Appendix~\ref{app:grader}.

\section{Outcome and Process Evaluation}
\label{app:grader}

This appendix details the evaluation procedure introduced
in \S\ref{sec:reward}.
Each retained execution receives separate outcome and
process verdicts.
We describe the outcome checker, the two process-grading
components, their composition, and the statistical
procedures used to summarize the recorded evaluations.

\subsection{Outcome Checker}
\label{app:outcome-checker}

The outcome checker determines whether the final business
state satisfies the task's expected result.
Its reference information comes from the customer goal
$g$ and the relevant private customer facts in the
task specification.
The check is domain-specific because the business records
and requested operations differ across domains.

For a banking dispute task, outcome correctness requires
a filed dispute whose relevant fields match the specified
customer facts, including the intended transaction.
The checker inspects the resulting business state;
it does not determine whether the agent provided the
required disclosures or followed the required sequence
of actions before filing.

The outcome verdict is binary:
\[
r_{\mathrm{outcome}} \in \{0,1\}.
\]
Two interactions that produce the same correct business
result can therefore receive the same outcome verdict
despite differing in process compliance.
These differences are assessed by the process grader.

\subsection{Deterministic Process Rules}
\label{app:deterministic-rules}

The deterministic rule layer evaluates the hard rules
specified for the task.
These checks include event ordering, tool permissions,
and agreement between confirmed and submitted fields.
The grading configuration identifies the applicable
hard rules through
\texttt{policy.\allowbreak grading\_\allowbreak policy.\allowbreak deterministic\_\allowbreak hard\_\allowbreak rules}.
The current grader enforces 53 hard rules
across the benchmark.

Each rule check produces a pass/fail result.
The deterministic component passes only when all
applicable hard checks pass:
\[
r_{\mathrm{rule}}
=
\bigwedge_{\text{applicable hard checks}}
\text{check result}.
\]
The rule outputs identify the checks responsible for
a failure and the associated recorded evidence.
Rule categories are described in Appendix~\ref{app:rules};
observed failure patterns are reported in
Appendix~\ref{app:rule-analysis}.

The grader validates referenced rule identifiers.
An unsupported identifier raises a grading error rather
than being silently recorded as a failed process rule.
This distinguishes an implementation or specification
problem from an observed violation by the agent.

For the submission-control experiment, both execution
modes use the original hybrid grading protocol
(\S\ref{sec:exp:control}).
Making submission tools available to the agent does
not itself remove the direct-submit prohibition from
that protocol.
The diagnostic that removes this prohibition is
reported separately.

\subsection{LLM Trace Judge}
\label{app:trace-judge}

The LLM judge assesses semantic process requirements
using an interpretation rubric and a task-specific
checklist.
The baseline evaluation uses \code{gpt-5.6-sol};
model and execution settings appear in
Appendix~\ref{app:repro}.

The judge receives the task context and grading checklist
alongside the recorded execution evidence.
The checklist is supplied through
\texttt{policy.llm\_judge.checklist}.
It specifies the requirements to be assessed, including
semantic conditions such as required disclosure content
and restrictions on unsupported outcome promises.

The judge returns a structured response with three
top-level fields:
\texttt{process\_pass}, \texttt{findings}, and
\texttt{rationale}.
Each finding contains:
\begin{itemize}
    \item \texttt{rule\_id}: the assessed requirement;
    \item \texttt{passed}: its pass/fail judgment;
    \item \texttt{severity}: \texttt{critical},
          \texttt{major}, or \texttt{minor};
    \item \texttt{seq}: a trace-event sequence reference;
    \item \texttt{evidence}: the evidence supporting
          the judgment.
\end{itemize}

The overall judge verdict is reconciled with these
individual findings before composition with the
deterministic component.
Any failed finding, regardless of severity, forces
the judge component to fail.
Expert review of the prompt and recorded judgments is
described in Appendix~\ref{app:validity}; sensitivity
to judge configuration is reported in
Appendix~\ref{app:robustness}.

\subsection{Verdict Composition and Consistency}
\label{app:verdict-composition}

The findings-consistency procedure reconciles the
judge's overall verdict with its individual findings.
For a schema-conforming judge response, it applies
the following rules:
\begin{itemize}
    \item If any finding fails, the reconciled judge
          verdict is fail.
    \item If the findings list is non-empty and every
          finding passes, the reconciled judge verdict
          is pass.
    \item If the findings list is empty, the original
          overall verdict is retained. In particular,
          an empty list does not overturn a fail verdict.
\end{itemize}

Any override is recorded with the grade.
This procedure addresses inconsistencies between the
overall verdict and the listed findings; it does not
independently establish that the findings correctly
interpret the trace.

The hybrid process verdict requires both components
to pass:
\[
r_{\mathrm{process}}
=
r_{\mathrm{rule}} \wedge r_{\mathrm{judge}}.
\]
Overall success and false-pass are then
\[
\begin{aligned}
r_{\mathrm{success}}
&=
r_{\mathrm{outcome}} \wedge r_{\mathrm{process}},\\
r_{\mathit{false\_pass}}
&=
r_{\mathrm{outcome}} \wedge
\neg r_{\mathrm{process}}.
\end{aligned}
\]
Rule-only and LLM-only evaluations are ablations;
the main results and submission-control comparison
use the hybrid grader.

Listing~\ref{lst:grade-run} summarizes the procedure.
The pseudocode shows the verdict logic; domain-specific
check implementations and override bookkeeping are
omitted.

\begin{lstlisting}[
    language=Python,
    caption={Outcome checking and hybrid process grading
    with verdict--findings consistency.},
    label={lst:grade-run}
]
def grade_run(task, trace, final_state, mode, model):
    validate_rule_ids(task.rules)

    hard_ids = (
        task.policy.grading_policy.deterministic_hard_rules
    )
    validate_rule_ids(hard_ids)

    r_outcome = outcome_pass(task, final_state)

    rule_results = [
        grade_rule(rid, task, trace, final_state, mode)
        for rid in hard_ids
    ]
    r_rule = all(result.passed for result in rule_results)

    judge_result = LLMTraceJudge(
        task, mode, model
    ).judge(trace, final_state)

    judge_result = enforce_findings_consistency(
        judge_result
    )
    r_judge = judge_result["process_pass"]

    r_process = r_rule and r_judge

    return {
        "outcome_pass": r_outcome,
        "process_pass": r_process,
        "rule_results": rule_results,
        "judge_result": judge_result,
    }
\end{lstlisting}

\subsection{Metric Definitions and Statistical Procedures}
\label{app:metrics}

\textbf{Evaluation unit and denominator.}
Let $N$ be the number of retained evaluation records
in the set being summarized.
The main evaluation retains one execution per model--task pair.
Any retries during data collection were triggered exclusively
by network failures, rather than by outcome or process verdicts.
The retry and retention procedure is described in
Appendix~\ref{app:evaluation-setup}.
Thus, $N=120$ for each model and $N=840$ for the
pooled seven-model evaluation.
These records do not represent every execution attempt
made during data collection.

All retained records are included in the reported
main rates.
The denominator is not restricted to executions that
terminate naturally without an error or budget limit.
The metrics therefore describe the retained evaluations,
rather than first-attempt reliability or variability
across repeated executions.

For each retained record $i$, let
\[
o_i := r_{\mathrm{outcome}}^{(i)},
\qquad
p_i := r_{\mathrm{process}}^{(i)}.
\]
Define the both-pass and false-pass indicators as
\[
b_i := o_i p_i,
\qquad
f_i := o_i(1-p_i).
\]

\textbf{Counts and all-record rates.}
The outcome-pass, both-pass, and false-pass counts are
\[
O := \sum_{i=1}^{N} o_i,
\qquad
B := \sum_{i=1}^{N} b_i,
\qquad
F := \sum_{i=1}^{N} f_i.
\]
Since every outcome-passing record either passes or
fails the process check,
\[
O = B + F.
\]
The corresponding rates over all retained records are
\[
\begin{aligned}
\widehat{R}_{\mathrm{outcome}} &= \frac{O}{N}, \\[2pt]
\widehat{R}_{\mathrm{both}} &= \frac{B}{N}, \\[2pt]
\widehat{R}_{\mathrm{false\text{-}pass}} &= \frac{F}{N}.
\end{aligned}
\]
Reported percentages multiply these proportions by 100.

\textbf{False-passes among outcome-passing records.}
To quantify the fraction of outcome-only successes
that fail the process check, we additionally report
\[
\widehat{R}_{\mathrm{false\text{-}pass}
\mid \mathrm{outcome}}
=
\frac{F}{O},
\qquad O>0.
\]
This conditional proportion has a different denominator
from the false-pass rate over all retained records.
In the main evaluation,
\[
\begin{aligned}
N &= 840, &\qquad O &= 582, \\[2pt]
B &= 116, &\qquad F &= 466.
\end{aligned}
\]
Thus, false-passes account for $466/840=55.48\%$
of all retained evaluations and
$466/582=80.07\%$ of outcome-passing evaluations.

\textbf{Confidence intervals.}
For each model, the reported 95\% confidence interval
applies to its both-pass proportion
$\widehat{p}=B/N$, with $N=120$.
We use the Wilson score interval
\[
\frac{
    \widehat{p}+\dfrac{z^2}{2N}
    \;\pm\;
    z\sqrt{
        \dfrac{\widehat{p}(1-\widehat{p})}{N}
        +\dfrac{z^2}{4N^2}
    }
}{
    1+\dfrac{z^2}{N}
},
\quad z=1.96.
\]
The minus and plus signs give the lower and upper
bounds, respectively.
The table reports both bounds as percentages.

\section{Benchmark Construction}
\label{app:construction}

The construction pipeline combines author-defined domain
components and regulatory requirements with LLM-assisted
task synthesis, automated structural validation, and
quality review.
Its outputs are task specifications and associated grading
criteria for \NTasks{} tasks across \NDomains{} domains,
with \NTasksPerDomain{} tasks per domain.
This appendix describes the inputs, responsibilities,
outputs, and checks involved in that process.

\subsection{Domain and Rule Preparation}
\label{app:domain-preparation}

\textbf{Inputs and responsibility.}
The authors use regulatory materials available through
the Legal Information Institute to identify obligations
relevant to the benchmark's customer-service workflows.
Domain preparation also requires defining the business
records and operations needed to represent those workflows.

\textbf{Outputs.}
We manually extracted 55 obligations and deduplicated
them into \NRules{} rule cards.
Each card records a stable identifier, source citation,
obligation, required trace evidence, and an associated
deterministic check or LLM-judge checklist item
(Appendix~\ref{app:rule-cards}).

The authors also define the domain's business-state
representation and implement its tool interface.
Obligations are mapped to existing trace predicates,
with additional predicates implemented where the
grader does not already support the required check.
These domain components provide the registered tools
and rule identifiers referenced during task synthesis.

The resulting preparation work therefore includes
regulatory interpretation, domain implementation,
and the connection between requirements and grading
criteria.
The shared runtime and grading framework provide
the common mechanisms used with these components.

\subsection{LLM-Assisted Task Synthesis}
\label{app:task-synthesis}

\textbf{Inputs and responsibility.}
LLM workers receive rule cards and a domain interface
manifest exposing the registered tools, their authority
levels, and the supported rule catalog.
The manifest supplies the operations and checks available
for constructing candidate tasks.

\textbf{Outputs.}
The workers generate candidate task specifications
and their grading specifications.
Each candidate instantiates a customer request in a
domain-specific business environment and associates
it with the relevant process requirements.
The task specification includes the customer goal,
opening message, initial business records, private
customer facts, tool manifest, and confirmation fields.
The grading specification identifies the applicable
deterministic rules and semantic checklist items.

The generated tasks use the YAML representation in
Appendix~\ref{app:taskspec}.
Rule cards provide a common reference for the scenario
and its evaluation criteria, while the interface
manifest constrains references to the benchmark's
supported operations and checks.

\subsection{Structural Validation and Quality Review}
\label{app:task-validation}

\textbf{Structural checks.}
Generated task specifications are checked automatically
against the task schema, domain registry, and rule catalog.
The pipeline applies four structural gates:
\begin{enumerate}
    \item Every field in
          \texttt{confirmation\_contract}
          appears in \texttt{hidden\_facts}.
    \item Every referenced submission tool is registered
          in the corresponding domain.
    \item Every referenced rule identifier is supported
          by the grader's rule catalog.
    \item The task provides a non-empty
          \texttt{llm\_judge.checklist}.
\end{enumerate}

A candidate that violates one of these conditions fails
structural validation.
For example, a task fails the confirmation-field check
if its confirmation contract names a transaction
identifier absent from the private customer facts.
These gates check the consistency of the specification
with the information and implementations available
to execute and grade it.

\textbf{Industry input and rule operationalization.}
Industry collaborators contributed business scenarios and
interview feedback, reviewed the encoded process rules,
and annotated execution traces.
The authors translated requirements from public regulatory
materials into scenario-specific rules, using institutional
customer-service procedures as workflow references.
The authors retain responsibility for the interpretation
of regulatory sources and their mapping to grading criteria.

Three senior software-engineering experts reviewed each rule
mapping for executable operationalization.
Their review covered the required behavior, observable trace
evidence, and timing or ordering constraints encoded in the
grading criteria.
Where reviewers disagreed, the disagreements were resolved
through discussion and voting.
The same three software-engineering experts also reviewed
the judging instructions.

These activities complement the automated structural checks.
Registered tools, supported rule identifiers, and complete
fields establish specification consistency.
Industry input informs the business context and workflows,
while engineering review examines their executable
representation and the judging instructions.
These activities do not constitute independent legal
validation of the encoded requirements.

\textbf{Review of the judge.}
Five senior software-engineering experts contributed to judge
development: one drafted the judging instructions, three reviewed
them, and a fifth calibrated the judge using 120 traces generated
by \code{gpt-5.4-mini}. Appendix~\ref{app:validity} describes
the procedure and reports the comparison of \code{gpt-6-astra}
judgments with the expert labels on this calibration set.

\subsection{Domain Extension and Construction Guidance}
\label{app:domain-extension}

Adding a domain requires the following work:
\begin{enumerate}
    \item Extract rule cards from the domain's
          regulatory requirements.
    \item Define the business-state representation
          and implement the domain tool interface.
    \item Map obligations to existing trace predicates
          or implement additional predicates where
          required.
    \item Generate and validate task specifications
          using the resulting rule cards, interfaces,
          and grading criteria.
\end{enumerate}

Domain-specific state representations, tool implementations,
and previously unsupported checks require additional
engineering.
The interaction loop, common trace format, submission
controls, verdict composition, and judge protocol are
reused across domains.
Task-defined confirmation fields connect different
business workflows to the shared submission mechanism
(Appendix~\ref{app:runtime}).

This process provides guidance for constructing further
domains or related benchmarks.
Supported operations and required trace evidence should
be established before generating scenarios, with tasks
and grading criteria derived from the same rule cards.
Structural validation checks whether the resulting
specifications are internally consistent and reference
supported components.
Semantic review separately examines whether the scenarios
and grading criteria preserve the meaning and applicability
of the source requirements.

\section{Experimental Setup and Reproducibility}
\label{app:repro}

\subsection{Evaluation Setup}
\label{app:evaluation-setup}

We evaluate seven agent models on \NTasks{} tasks across
\NDomains{} domains, with \NTasksPerDomain{} tasks per domain.
The models are listed in Table~\ref{tab:main}.
The analysis includes one retained execution per model--task
pair, yielding 840 evaluation records.
Retries during data collection were triggered exclusively
by network failures. Outcome failure, process violations,
and exhaustion of the dialogue or tool-call budget were
not retry triggers.
All 120 retained records per model, including task failures,
are included in the reported rates.

The customer simulator is \code{gpt-5.4-mini}, and the
baseline judge is \code{gpt-5.6-sol}.
Unless otherwise specified, generation parameters,
including temperature, and dialogue and tool-call
budgets use their default settings.

The main evaluation uses mediated submission.
The submission-control experiment compares mediated
and direct submission for \code{gpt-5.4} on the same
120 tasks under the original hybrid grading protocol
(\S\ref{sec:exp:control}).
Metric definitions and statistical procedures appear
in Appendix~\ref{app:metrics}.

\subsection{Judge-Configuration Comparison}
\label{app:judge-comparison-setup}

We compare \code{gpt-5.6-sol} and \code{gpt-6-astra}
on the same 720 recorded traces from all evaluated
agent models except \code{deepseek-v4-flash}.
Both judges use an output-token limit of 81,920.

The comparison holds the recorded interactions fixed
and examines changes in process verdicts, aggregate
scores, and model ordering.
We cannot verify whether provider-side system instructions
are identical across the two judge models, limiting
attribution of observed differences solely to
judge-model choice.
Comparison results are reported in
Appendix~\ref{app:robustness}.

\subsection{Reproducing the Results}
\label{app:reproduction-procedures}

We distinguish three reproduction procedures.

\textbf{Task re-execution.}
Running a task again invokes the agent and customer
simulator through the runtime, producing a new
interaction whose trace and verdicts may differ
from the retained execution.

\textbf{Trace re-grading.}
Re-grading applies a specified grading configuration
to the recorded interaction and final business state.
The execution evidence remains fixed, while fresh
LLM judgments may produce different findings.
The judge-configuration comparison follows this
procedure.

\textbf{Statistical reaggregation.}
Recomputing tables uses saved outcome and process
verdicts without additional model calls.
For the same retained records, grades, and statistical
procedures, the reported counts, rates, confidence
intervals, and paired comparisons are deterministically
recomputed.

\subsection{Data Availability}
\label{app:data-availability}
\label{sec:data-availability}

The reproducibility artifact contains the task corpus, domain tools, shared runtime,
rule grader, trace-judge protocol, and task-construction
pipeline, together with the recorded run logs and
grading records supporting the results.

The artifact includes a container image and scripts
for regenerating the reported tables from saved
evaluations.
Secrets, account metadata, and raw provider payloads
are excluded.
The released materials support inspection of the
requirements, execution evidence, and grading decisions
underlying the reported results.

\section{Additional Results}
\label{app:results}

This section provides domain-level results, rule-failure
analyses, submission-control results, and an annotated
false-pass example.
Unless otherwise stated, results use the 840 retained
executions from seven models on 120 tasks under
mediated-submit execution.
All rates include failed tasks, and process verdicts
use the hybrid grader.

\subsection{Per-Domain Results}
\label{app:per-domain}

\begin{figure*}[t]
    \centering
    \includegraphics[width=\textwidth]
        {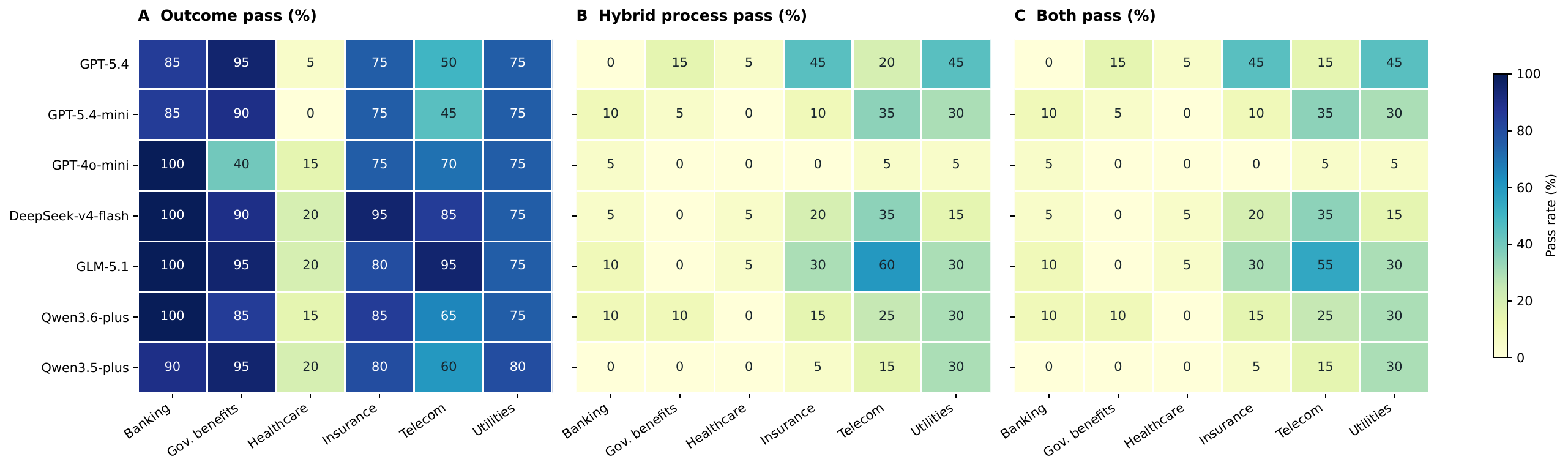}
    \caption{Outcome, hybrid process, and both-pass rates
    across models and domains.
    Each cell uses all 20 tasks for one model--domain pair.
    All panels share a 0--100\% color scale.}
    \label{fig:domain-heatmap-app}
\end{figure*}

Figure~\ref{fig:domain-heatmap-app} reports outcome,
process, and both-pass rates for each model--domain pair.
Table~\ref{tab:domain-detail-app} aggregates the seven
models within each domain.

Banking and government benefits show particularly large
outcome--process gaps: outcome-pass reaches 94.29\%
and 84.29\%, respectively, whereas both-pass is only
5.71\% and 4.29\%.
Healthcare has the lowest pooled both-pass rate at 2.14\%.
Telecommunications and utilities share the highest
both-pass rate at 26.43\%, although both retain
substantial false-pass rates.

\begin{table}[t]
    \centering
    \caption{Pooled results by domain.
    Each domain contains 140 evaluations.
    All entries are percentages.}
    \label{tab:domain-detail-app}
    \footnotesize
    \setlength{\tabcolsep}{3pt}
    \renewcommand{\arraystretch}{1.1}
    \begin{tabular*}{\linewidth}{
        @{\extracolsep{\fill}}lrrr@{}
    }
        \toprule
        Domain & \shortstack[r]{Outcome\\pass}
               & \shortstack[r]{Both\\pass}
               & \shortstack[r]{False\\pass} \\
        \midrule
        Banking            & 94.29 & 5.71  & 88.57 \\
        Government benefits & 84.29 & 4.29  & 80.00 \\
        Healthcare         & 13.57 & 2.14  & 11.43 \\
        Insurance          & 80.71 & 17.86 & 62.86 \\
        Telecommunications & 67.14 & 26.43 & 40.71 \\
        Utilities          & 75.71 & 26.43 & 49.29 \\
        \bottomrule
    \end{tabular*}
\end{table}

\subsection{Rule-Level Failure Analysis}
\label{app:rule-analysis}

\begin{figure*}[t]
    \centering
    \includegraphics[width=\textwidth]
        {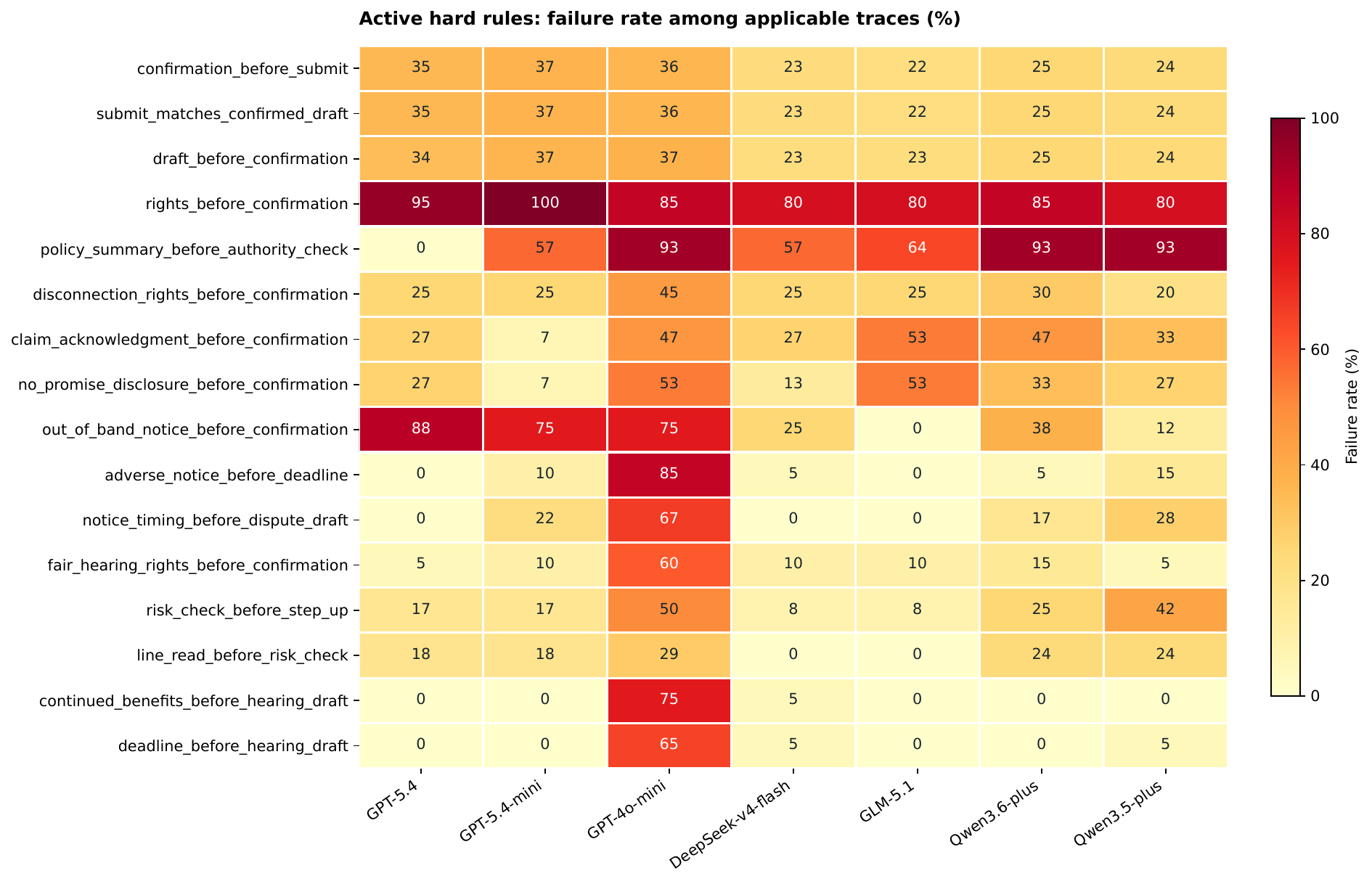}
    \caption{Failure rates for the 16 most frequently
    failing hard rules across the seven models.
    Each rate is normalized by the number of records
    whose task configures that rule for the corresponding
    model, so denominators vary by rule.
    Failures include missing required events as well as
    incorrect event ordering.}
    \label{fig:rule-failure-app}
\end{figure*}

Of the 53 hard rules used in the evaluated tasks,
35 fail at least once.
Figure~\ref{fig:rule-failure-app} shows the failure rates
of the 16 most frequently failing hard rules.
Each rate uses the records whose task configures the
corresponding rule as its denominator.

A recurring insurance failure concerns
\texttt{policy\_\allowbreak{}summary\_\allowbreak{}before\_%
\allowbreak{}authority\_\allowbreak{}check},
an ordering requirement derived from the referenced
institutional workflow. It requires a successful
policy-summary read before the reporter-authority check.
This rule fails in 47 false-pass records, accounting for
10.09\% of all 466 false-passes.

This proportion describes records in which the workflow
requirement fails; it does not attribute those false-passes
exclusively to this rule, because other requirements may
also fail. Moreover, the check can fail because a required
event is missing or because the events occur in the
wrong order.

Both grader components contribute to identifying
false-passes.
Of the 466 false-passes, 329 are rejected only by the
LLM judge, 38 only by hard rules, and 99 by both.
These counts show that the two components detect
different process failures accompanying correct outcomes.

\subsection{Submission-Control Ablation}
\label{app:direct-submit}

We compare \code{gpt-5.4} under mediated-submit and
direct-submit using the same 120 task specifications and recorded
traces. The agent and customer simulator were not rerun. Under
the original hybrid protocol, both-pass falls from 20.83\% to
0.83\% (two-sided exact McNemar
$p=1.19\times10^{-7}$), while false-pass rises from 43.33\% to
66.67\%. Outcome-pass changes from 64.17\% to 67.50\%
($p=0.4545$).

The original protocol includes submission-mechanism checks that
specifically penalize direct submission, including
\texttt{no\_agent\_direct\_submit},
\texttt{confirmation\_before\_submit}, and
\texttt{submit\_matches\_confirmed\_draft}, together with
corresponding LLM-judge subconditions. We therefore report a
third scoring condition, \emph{Direct-submit (submission-mechanism
checks excluded)}. This condition rescored the same traces after
excluding these checks from both grading components while
retaining all other hard rules and semantic findings. It is a
scoring condition rather than an additional execution mode.

Under the same excluded-check scoring applied to both modes,
mediated-submit remains at 25/120 both-pass runs (20.83\%), while
direct-submit reaches 22/120 (18.33\%). The direct-minus-mediated
difference is $-2.50$ percentage points, with two-sided exact
McNemar $p=0.677639$ and a paired-task bootstrap 95\% interval of
$[-10.00,5.00]$ percentage points. The corresponding
direct-submit process-pass and false-pass rates are 20.00\% and
49.17\%, respectively.

For the excluded-check comparison, 12 tasks pass under both modes,
13 pass only under mediated-submit, 10 pass only under
direct-submit, and 85 fail under both. As an intermediate
diagnostic, removing only
\texttt{no\_agent\_direct\_submit} yields 9/120 direct-submit
both-pass runs; this condition is not used for the primary
comparison because the other submission-mechanism checks remain
active.

\begin{table}[t]
    \centering
    \caption{Submission-control ablation for
    \code{gpt-5.4}. All rows use the same 120 task specifications
    and recorded traces. The third row excludes
    submission-mechanism checks from both the hard-rule and
    LLM-judge components; it is a rescoring condition rather than
    an additional execution mode. All entries are percentages.}
    \label{tab:direct-submit-app}
    \footnotesize
    \setlength{\tabcolsep}{2pt}
    \renewcommand{\arraystretch}{1.1}
    \begin{tabular*}{\linewidth}{
        @{\extracolsep{\fill}}
        >{\raggedright\arraybackslash}p{0.30\linewidth}
        rrrr@{}
    }
        \toprule
        Mode
        & \shortstack[r]{Outcome\\pass}
        & \shortstack[r]{Process\\pass}
        & \shortstack[r]{Both\\pass}
        & \shortstack[r]{False\\pass} \\
        \midrule
        Mediated-submit
        & 64.17 & 21.67 & 20.83 & 43.33 \\
        Direct-submit
        & 67.50 & 0.83 & 0.83 & 66.67 \\
        Direct-submit\newline (checks excluded)
        & 67.50 & 20.00 & 18.33 & 49.17 \\
        \bottomrule
    \end{tabular*}
\end{table}

\subsection{Annotated False-Pass Trace}
\label{app:trace}

The \code{gpt-5.4-mini} execution of insurance task
\texttt{INS-FLOW-010.us} illustrates a correct outcome
accompanied by process failures.
Listing~\ref{lst:false-pass-app} summarizes selected
events under mediated-submit execution, retaining
their original sequence numbers.

\begin{lstlisting}[
    language={},
    numbers=none,
    basicstyle=\ttfamily\scriptsize,
    breaklines=true,
    columns=fullflexible,
    caption={A confirmed total-loss-review draft is
    submitted successfully despite unmet process
    requirements.},
    label={lst:false-pass-app}
]
seq 10--11, agent/tool:
  Reads the policy summary.

seq 17--18, agent/tool:
  Checks total-loss or storage risk.

seq 20--22, agent/runtime/tool:
  Creates draft_001 for a total-loss review.

seq 24--28, agent/tool:
  Records acknowledgment and no-promise notices
  for draft_001.

seq 34--36, agent/runtime/tool:
  Creates corrected draft_002.

seq 39--41, simulator/customer/runtime:
  Confirms draft_002 and submits the matching draft.
\end{lstlisting}

The submitted review matches the required task fields,
so $r_{\mathrm{outcome}}=1$.
However, the deterministic rule layer identifies a
missing reporter-authority check.
The LLM judge additionally flags missing existing-claim
context before the business check and insufficient
association or carry-forward of the required notices
after the draft is corrected.

Both grader components therefore fail:
$r_{\mathrm{rule}}=0$ and $r_{\mathrm{judge}}=0$,
making this execution a false-pass.
The example illustrates how matching draft confirmation
and successful submission can coexist with unmet
process requirements.

\section{Evaluation Sensitivity and Expert Review}
\label{app:robustness}

We examine sensitivity to judge configuration and hard-rule
selection, report the effect of verdict reconciliation,
and summarize expert review of the judge prompt and
trace-level judgments.
\subsection{Comparison of Judge Configurations}
\label{app:judge-comparison}

We compare \code{gpt-5.6-sol} and \code{gpt-6-astra}
on the same 720 traces from six models, excluding
\code{deepseek-v4-flash}.
Outcome checks and deterministic rule results remain fixed.
Both judges use an output token limit of 81,920;
the evaluation settings are summarized in
Appendix~\ref{app:repro}.

Table~\ref{tab:judge-substitution-app} reports the resulting
hybrid both-pass rates.
The pooled both-pass rate changes from 13.89\% to 14.72\%,
an increase of 0.83 percentage points.
Hybrid process verdicts agree on 92.22\% of evaluations.
The model ordering remains largely unchanged, with
Kendall's $\tau_b=0.9661$.
The only ordering change resolves the baseline tie between
\code{gpt-5.4-mini} and \code{qwen3.6-plus};
no previously strict pairwise ordering reverses.
These results indicate similar aggregate scores and model
ordering across the tested configurations, while agreement
measures consistency rather than correctness.

\begin{table}[t]
    \centering
    \caption{Both-pass rates under two judge configurations.
    Each model uses the same 120 traces for both judges.
    Rates are percentages; $\Delta$ is the change from
    \code{gpt-5.6-sol} to \code{gpt-6-astra}
    in percentage points.}
    \label{tab:judge-substitution-app}
    \footnotesize
    \setlength{\tabcolsep}{1.5pt}
    \renewcommand{\arraystretch}{1.1}
    \begin{tabular*}{\linewidth}{
        @{\extracolsep{\fill}}lrrr@{}
    }
        \toprule
        Agent
        & \code{gpt-5.6-sol}
        & \code{gpt-6-astra}
        & $\Delta$ (pp) \\
        \midrule
        \code{gpt-5.4}
        & 20.83 & 20.00 & $-0.83$ \\
        \code{gpt-5.4-mini}
        & 15.00 & 17.50 & $+2.50$ \\
        \code{gpt-4o-mini}
        & 2.50 & 2.50 & $0.00$ \\
        \code{glm-5.1}
        & 21.67 & 28.33 & $+6.67$ \\
        \code{qwen3.6-plus}
        & 15.00 & 14.17 & $-0.83$ \\
        \code{qwen3.5-plus}
        & 8.33 & 5.83 & $-2.50$ \\
        \bottomrule
    \end{tabular*}
\end{table}

\subsection{Sensitivity to Individual Hard Rules}
\label{app:rule-sensitivity}

We remove each of the 35 hard rules with observed failures
in turn, holding outcomes, recorded judge verdicts, and
the remaining hard rules fixed.
Among the six models other than \code{gpt-4o-mini},
only three rule removals increase any model's both-pass count.

The largest increase occurs when removing
\texttt{policy\_\allowbreak{}summary\_\allowbreak{}before\_%
\allowbreak{}authority\_\allowbreak{}check}:
\code{qwen3.6-plus} rises from 18 to 24 both-passes,
an increase of 5.00 percentage points.
Removing either
\texttt{read\_\allowbreak{}before\_\allowbreak{}check}
or
\texttt{risk\_\allowbreak{}check\_\allowbreak{}before\_%
\allowbreak{}step\_\allowbreak{}up}
changes a model's rate by at most 0.83 percentage points.

Two rule removals break the baseline tie between
\code{gpt-5.4-mini} and \code{qwen3.6-plus},
but none reverses a previously strict ordering.
These analyses measure sensitivity to individual rules;
the reported benchmark scores retain the full hybrid grader.

\subsection{Effect of Verdict Reconciliation}
\label{app:consistency-analysis}

We examine the effect of the verdict--findings consistency
check described in Appendix~\ref{app:grader}.
In the 840 baseline evaluations, this check changes two
raw judge verdicts, both for \code{gpt-5.4-mini}.
Both records already fail outcome and hard-rule evaluation,
so the both-pass count remains 116 and the false-pass
count remains 466.

The check changes no judge verdicts in the 720
alternative-judge evaluations or the 120 direct-submit
evaluations.
It resolves internal inconsistencies between a judge's
verdict and findings; it does not independently verify
the interpretation of trace evidence.
\subsection{Expert Review and Human-Reference Comparison}
\label{app:validity}

\textbf{Participants and development procedure.}
Industry collaborators contributed business scenarios and
interview feedback, reviewed the encoded process rules,
and annotated execution traces.
The authors were responsible for extracting requirements
from regulatory sources and translating them into rule cards
and grading criteria. Five senior
software-engineering experts contributed to judge development.
One expert drafted the initial judging instructions, and three
additional experts reviewed them. Disagreements during this
review were resolved through discussion and voting. After the
initial runs, a fifth expert reviewed 120 traces generated by
the \code{gpt-5.4-mini} agent and compared their trace-level
judgments with \code{gpt-5.6-sol} verdicts to refine the judge.
The trace-level reference judgments for these 120 calibration
traces were produced by the calibration expert described above.

\textbf{Comparison with expert labels.}
Table~\ref{tab:human-reference} compares \code{gpt-5.6-sol}
verdicts with the expert labels on these 120 traces. With process
pass as the positive class, there are 23 true positives, 4 false
negatives, 10 false positives, and 83 true negatives. Precision
is $23/33=69.70\%$, recall is $23/27=85.19\%$, F1 is
$46/60=76.67\%$, and accuracy is $106/120=88.33\%$.

With process failure as the positive class, precision is
$83/87=95.40\%$ and recall is $83/93=89.25\%$.
Thus, the expert supports 83 of the judge's 87 failure verdicts;
4 expert-labeled passes are rejected, and 10 expert-labeled
failures are accepted.

\begin{table}[t]
    \centering
    \caption{Comparison of \code{gpt-5.6-sol} verdicts with
    expert labels on 120 traces generated by \code{gpt-5.4-mini}.
    These traces were used during judge calibration.
    Process pass is the positive class.}
    \label{tab:human-reference}
    \footnotesize
    \setlength{\tabcolsep}{3pt}
    \begin{tabular}{@{}lrr@{}}
        \toprule
        & \code{gpt-5.6-sol}: Pass
        & \code{gpt-5.6-sol}: Fail \\
        \midrule
        Expert: Pass & 23 (TP) & 4 (FN) \\
        Expert: Fail & 10 (FP) & 83 (TN) \\
        \bottomrule
    \end{tabular}
\end{table}

\textbf{Scope of the evidence.}
These traces were used during judge development and therefore
do not form an independent held-out validation set. The comparison
uses one expert's trace-level judgments and does not measure
inter-rater reliability. It evaluates the baseline
\code{gpt-5.6-sol} judge on traces from one agent model and does
not establish its accuracy across all evaluated models.
Industry participation provides practical input into the
benchmark's scenarios, workflows, rules, and trace annotations.
The source citations make the authors' regulatory
interpretations and rule mappings inspectable.
Neither industry participation nor source traceability
establishes the legal correctness or completeness of those
interpretations, or certifies compliance in deployment.
The 92.22\% hybrid-verdict agreement on the 720-trace
judge-comparison set measures consistency between configurations,
rather than accuracy against human labels.

%% file: arxiv.bbl
\begin{thebibliography}{17}
\providecommand{\natexlab}[1]{#1}

\bibitem[{Barres et~al.(2025)Barres, Dong, Ray, Si, and Narasimhan}]{barres2025tau2benchevaluatingconversationalagents}
Victor Barres, Honghua Dong, Soham Ray, Xujie Si, and Karthik Narasimhan. 2025.
\newblock \href {https://arxiv.org/abs/2506.07982} {$\tau^2$-bench: Evaluating conversational agents in a dual-control environment}.
\newblock \emph{Preprint}, arXiv:2506.07982.

\bibitem[{{Consumer Financial Protection Bureau}(2026)}]{cfpb2026rege}
{Consumer Financial Protection Bureau}. 2026.
\newblock \href {https://www.consumerfinance.gov/rules-policy/regulations/1005/11} {12 {CFR} 1005.11: Procedures for resolving errors}.
\newblock Electronic Code of Federal Regulations, Regulation E.
\newblock See paragraph (b)(2).

\bibitem[{Gogani-Khiabani et~al.(2025)Gogani-Khiabani, Trivedi, Saha, and Tizpaz-Niari}]{gogani2025legalcriticalagentic}
Sina Gogani-Khiabani, Ashutosh Trivedi, Diptikalyan Saha, and Saeid Tizpaz-Niari. 2025.
\newblock \href {https://arxiv.org/abs/2509.13471} {An {LLM} agentic approach for legal-critical software: A case study for tax prep software}.
\newblock \emph{Preprint}, arXiv:2509.13471.

\bibitem[{Guha et~al.(2023)Guha, Nyarko, Ho, Re, Chilton, Narayana, Chohlas-Wood, Peters, Waldon, Rockmore, Zambrano, Talisman, Hoque, Surani, Fagan, Sarfaty, Dickinson, Porat, Hegland, Wu, Nudell, Niklaus, Nay, Choi, Tobia, Hagan, Ma, Livermore, Rasumov-Rahe, Holzenberger, Kolt, Henderson, Rehaag, Goel, Gao, Williams, Gandhi, Zur, Iyer, and Li}]{guha2023legalbenchcollaborativelybuiltbenchmark}
Neel Guha, Julian Nyarko, Daniel~E. Ho, Christopher Re, Adam Chilton, Aditya Narayana, Alex Chohlas-Wood, Austin Peters, Brandon Waldon, Daniel~N. Rockmore, Diego Zambrano, Dmitry Talisman, Enam Hoque, Faiz Surani, Frank Fagan, Galit Sarfaty, Gregory~M. Dickinson, Haggai Porat, Jason Hegland, and 21 others. 2023.
\newblock \href {https://arxiv.org/abs/2308.11462} {Legalbench: A collaboratively built benchmark for measuring legal reasoning in large language models}.
\newblock \emph{Preprint}, arXiv:2308.11462.

\bibitem[{Jimenez et~al.(2024)Jimenez, Yang, Wettig, Yao, Pei, Press, and Narasimhan}]{jimenez2024swebenchlanguagemodelsresolve}
Carlos~E. Jimenez, John Yang, Alexander Wettig, Shunyu Yao, Kexin Pei, Ofir Press, and Karthik Narasimhan. 2024.
\newblock \href {https://arxiv.org/abs/2310.06770} {Swe-bench: Can language models resolve real-world github issues?}
\newblock \emph{Preprint}, arXiv:2310.06770.

\bibitem[{Li et~al.(2025)Li, Huang, Wang, Zhang, Antoniades, Hua, Zhu, Zeng, Wang, Wang, and Yan}]{li2025sopbench}
Zekun Li, Shinda Huang, Jiangtian Wang, Nathan Zhang, Antonis Antoniades, Wenyue Hua, Kaijie Zhu, Sirui Zeng, Chi Wang, William~Yang Wang, and Xifeng Yan. 2025.
\newblock \href {https://arxiv.org/abs/2503.08669} {{SOPBench}: Evaluating language agents at following standard operating procedures and constraints}.
\newblock \emph{Preprint}, arXiv:2503.08669.

\bibitem[{Liu et~al.(2024)Liu, Yu, Zhang, Xu, Lei, Lai, Gu, Ding, Men, Yang, Zhang, Deng, Zeng, Du, Zhang, Shen, Zhang, Su, Sun, Huang, Dong, and Tang}]{liu2025agentbenchevaluatingllmsagents}
Xiao Liu, Hao Yu, Hanchen Zhang, Yifan Xu, Xuanyu Lei, Hanyu Lai, Yu~Gu, Hangliang Ding, Kaiwen Men, Kejuan Yang, Shudan Zhang, Xiang Deng, Aohan Zeng, Zhengxiao Du, Chenhui Zhang, Sheng Shen, Tianjun Zhang, Yu~Su, Huan Sun, and 3 others. 2024.
\newblock \href {https://arxiv.org/abs/2308.03688} {Agentbench: Evaluating llms as agents}.
\newblock In \emph{Proceedings of the 12th International Conference on Learning Representations}.

\bibitem[{Liu et~al.(2023)Liu, Iter, Xu, Wang, Xu, and Zhu}]{liu2023gevalnlgevaluationusing}
Yang Liu, Dan Iter, Yichong Xu, Shuohang Wang, Ruochen Xu, and Chenguang Zhu. 2023.
\newblock \href {https://arxiv.org/abs/2303.16634} {G-eval: Nlg evaluation using gpt-4 with better human alignment}.
\newblock \emph{Preprint}, arXiv:2303.16634.

\bibitem[{Nandi et~al.(2025)Nandi, Datta, Nama, Patel, Vichare et~al.}]{nandi2025sopbench}
Subhrangshu Nandi, Arghya Datta, Rohith Nama, Udita Patel, Nikhil Vichare, et~al. 2025.
\newblock \href {https://arxiv.org/abs/2506.08119} {{SOP-Bench}: Complex industrial standard operating procedures for evaluating {LLM} agents}.
\newblock \emph{Preprint}, arXiv:2506.08119.

\bibitem[{{OWASP Agentic Security Initiative}(2025)}]{owasp2025agenticthreats}
{OWASP Agentic Security Initiative}. 2025.
\newblock \href {https://genai.owasp.org/resource/agentic-ai-threats-and-mitigations/} {Agentic {AI} -- threats and mitigations}.
\newblock OWASP GenAI Security Project.
\newblock Defines the \emph{least agency} principle for agentic applications.

\bibitem[{Wang et~al.(2025{\natexlab{a}})Wang, Poskitt, and Sun}]{wang2025agentspeccustomizableruntimeenforcement}
Haoyu Wang, Christopher~M. Poskitt, and Jun Sun. 2025{\natexlab{a}}.
\newblock \href {https://arxiv.org/abs/2503.18666} {Agentspec: Customizable runtime enforcement for safe and reliable llm agents}.
\newblock \emph{Preprint}, arXiv:2503.18666.
\newblock Accepted at ICSE 2026.

\bibitem[{Wang et~al.(2025{\natexlab{b}})Wang, Tang, Jin, Chen, Jin, Ding, Li, and Cao}]{wang2025sopmaze}
Jiaming Wang, Zhe Tang, Zehao Jin, Hefei Chen, Yilin Jin, Peng Ding, Xiaoyu Li, and Xuezhi Cao. 2025{\natexlab{b}}.
\newblock \href {https://arxiv.org/abs/2510.08942} {{SOP-Maze}: Evaluating large language models on complicated business standard operating procedures}.
\newblock \emph{Preprint}, arXiv:2510.08942.

\bibitem[{Xie et~al.(2024)Xie, Zhang, Chen, Li, Zhao, Cao, Hua, Cheng, Shin, Lei, Liu, Xu, Zhou, Savarese, Xiong, Zhong, and Yu}]{xie2024osworldbenchmarkingmultimodalagents}
Tianbao Xie, Danyang Zhang, Jixuan Chen, Xiaochuan Li, Siheng Zhao, Ruisheng Cao, Toh~Jing Hua, Zhoujun Cheng, Dongchan Shin, Fangyu Lei, Yitao Liu, Yiheng Xu, Shuyan Zhou, Silvio Savarese, Caiming Xiong, Victor Zhong, and Tao Yu. 2024.
\newblock \href {https://arxiv.org/abs/2404.07972} {Osworld: Benchmarking multimodal agents for open-ended tasks in real computer environments}.
\newblock \emph{Preprint}, arXiv:2404.07972.

\bibitem[{Xu et~al.(2024)Xu, Song, Li, Tang, Jain, Bao, Wang, Zhou, others, and Neubig}]{xu2024theagentcompany}
Frank~F. Xu, Yufan Song, Boxuan Li, Yuxuan Tang, Kritanjali Jain, Mengxue Bao, Zora~Z. Wang, Xuhui Zhou, others, and Graham Neubig. 2024.
\newblock \href {https://arxiv.org/abs/2412.14161} {{TheAgentCompany}: Benchmarking {LLM} agents on consequential real world tasks}.
\newblock \emph{Preprint}, arXiv:2412.14161.

\bibitem[{Xu et~al.(2025)Xu, Gao, He, and Zhou}]{xu2024licoevalevaluatingllmslicense}
Weiwei Xu, Kai Gao, Hao He, and Minghui Zhou. 2025.
\newblock \href {https://doi.org/10.1109/ICSE55347.2025.00052} {Licoeval: Evaluating llms on license compliance in code generation}.
\newblock In \emph{Proceedings of the 47th IEEE/ACM International Conference on Software Engineering}.

\bibitem[{Yao et~al.(2024)Yao, Shinn, Razavi, and Narasimhan}]{yao2024taubenchbenchmarktoolagentuserinteraction}
Shunyu Yao, Noah Shinn, Pedram Razavi, and Karthik Narasimhan. 2024.
\newblock \href {https://arxiv.org/abs/2406.12045} {$\tau$-bench: A benchmark for tool-agent-user interaction in real-world domains}.
\newblock \emph{Preprint}, arXiv:2406.12045.

\bibitem[{Zheng et~al.(2023)Zheng, Chiang, Sheng, Zhuang, Wu, Zhuang, Lin, Li, Li, Xing, Zhang, Gonzalez, and Stoica}]{zheng2023judgingllmasajudgemtbenchchatbot}
Lianmin Zheng, Wei-Lin Chiang, Ying Sheng, Siyuan Zhuang, Zhanghao Wu, Yonghao Zhuang, Zi~Lin, Zhuohan Li, Dacheng Li, Eric~P. Xing, Hao Zhang, Joseph~E. Gonzalez, and Ion Stoica. 2023.
\newblock \href {https://arxiv.org/abs/2306.05685} {Judging llm-as-a-judge with mt-bench and chatbot arena}.
\newblock \emph{Preprint}, arXiv:2306.05685.

\end{thebibliography}
